\PassOptionsToPackage{dvipsnames,table}{xcolor}
\documentclass[acmtog]{acmart}
\usepackage{algorithm2e}
\usepackage[inline]{enumitem}
\usepackage[dvipsnames,table]{xcolor}
\AtBeginDocument{%
  \providecommand\BibTeX{{%
    \normalfont B\kern-0.5em{\scshape i\kern-0.25em b}\kern-0.8em\TeX}}}

\copyrightyear{2024}
\acmYear{2024}
\setcopyright{acmlicensed}\acmConference[SA Conference Papers '24]{SIGGRAPH Asia 2024 Conference Papers}{December 3--6, 2024}{Tokyo, Japan}
\acmBooktitle{SIGGRAPH Asia 2024 Conference Papers (SA Conference Papers '24), December 3--6, 2024, Tokyo, Japan}
\acmDOI{10.1145/3680528.3687646}
\acmISBN{979-8-4007-1131-2/24/12}

\acmSubmissionID{papers\_722}

\begin{document}

\title{InfNeRF: Towards Infinite Scale NeRF Rendering with $\mathcal{O}(\log n)$ Space Complexity}

\author{Jiabin Liang}
\email{liangjb@sea.com}
\authornote{Corresponding author.}
\orcid{0009-0000-3087-4540}
\affiliation{%
  \institution{Sea AI Lab}
  \streetaddress{1 Fusionopolis Place, \#17-10, Galaxis}
  \city{Singapore}
  \state{Singapore}
  \country{Singapore}
  \postcode{138522}
}

\author{Lanqing Zhang}
\email{zhanglq@sea.com}
\orcid{0009-0006-4662-809X}
\affiliation{%
   \institution{Sea AI Lab}
  \streetaddress{1 Fusionopolis Place, \#17-10, Galaxis}
  \city{Singapore}
  \state{Singapore}
  \country{Singapore}
  \postcode{138522}
}

\author{Zhuoran Zhao}
\email{zhuoran.zhao@u.nus.edu}
\orcid{0009-0008-3944-2099}
\affiliation{
  \institution{National University of Singapore}
  \country{Singapore}
}

\author{Xiangyu Xu}
\email{xuxiangyu2014@gmail.com}
\orcid{0000-0002-9305-5830}
\affiliation{%
  \institution{Xi'an Jiaotong University}
  \city{Xi'an}
  \country{China}
  \postcode{710049}
}

\renewcommand{\shortauthors}{Liang, et al.}
%\newcommand{\xy}[1]{\textcolor{red}{XY: #1}}
%\newcommand{\zlq}[1]{\textcolor{blue}{LQ: #1}}
%\newcommand{\jb}[1]{\textcolor{magenta}{JB: #1}}

\begin{abstract}
  The conventional mesh-based Level of Detail (LoD) technique, exemplified by applications such as Google Earth and many game engines, exhibits the capability to holistically represent a large scene even the Earth, and achieves rendering with a space complexity of $\mathcal{O}(\log n)$.
  %extend more about the n and what is dynamic fetching? or local data dependency?
  %No matter rendering the whole scene from a bird's eye view nor rendering a small detail from a close up view, it only required caching small amount of the data locally. 
  This constrained data requirement not only enhances rendering efficiency but also facilitates dynamic data fetching, thereby enabling a seamless 3D navigation experience for users.
  
  In this work, we extend this proven LoD technique to Neural Radiance Fields (NeRF) by introducing an octree structure to represent the scenes in different scales. 
  This innovative approach provides a mathematically simple and elegant representation with a rendering space complexity of $\mathcal{O}(\log n)$, aligned with the efficiency of mesh-based LoD techniques.
  We also present a novel training strategy that maintains a complexity of $\mathcal{O}(n)$. 
  This strategy allows for parallel training with minimal overhead, ensuring the scalability and efficiency of our proposed method. 
  Our contribution is not only in extending the capabilities of existing techniques but also in establishing a foundation for scalable and efficient large-scale scene representation using NeRF and octree structures. Code and checkpoints are available at: \url{https://jiabinliang.github.io/InfNeRF.io/}
\end{abstract}

\begin{CCSXML}
<ccs2012>
   <concept>
       <concept_id>10010147.10010178.10010224.10010245.10010254</concept_id>
       <concept_desc>Computing methodologies~Reconstruction</concept_desc>
       <concept_significance>500</concept_significance>
       </concept>
   <concept>
       <concept_id>10010147.10010371.10010372</concept_id>
       <concept_desc>Computing methodologies~Rendering</concept_desc>
       <concept_significance>500</concept_significance>
       </concept>
    <concept>
       <concept_id>10010147.10010371.10010396.10010401</concept_id>
       <concept_desc>Computing methodologies~Volumetric models</concept_desc>
       <concept_significance>300</concept_significance>
       </concept>
   <concept>
       <concept_id>10010147.10010371.10010382.10010386</concept_id>
       <concept_desc>Computing methodologies~Antialiasing</concept_desc>
       <concept_significance>300</concept_significance>
       </concept>
 </ccs2012>
\end{CCSXML}
\ccsdesc[500]{Computing methodologies~Reconstruction}
\ccsdesc[500]{Computing methodologies~Rendering}
\ccsdesc[300]{Computing methodologies~Volumetric models}
\ccsdesc[300]{Computing methodologies~Antialiasing}

\keywords{novel view synthesis, radiance fields, level of detail, }

%\received{20 February 2024}
%\received[revised]{12 March 2024}
%\received[accepted]{5 June 2024}
\begin{teaserfigure}
  \includegraphics[width=\textwidth]{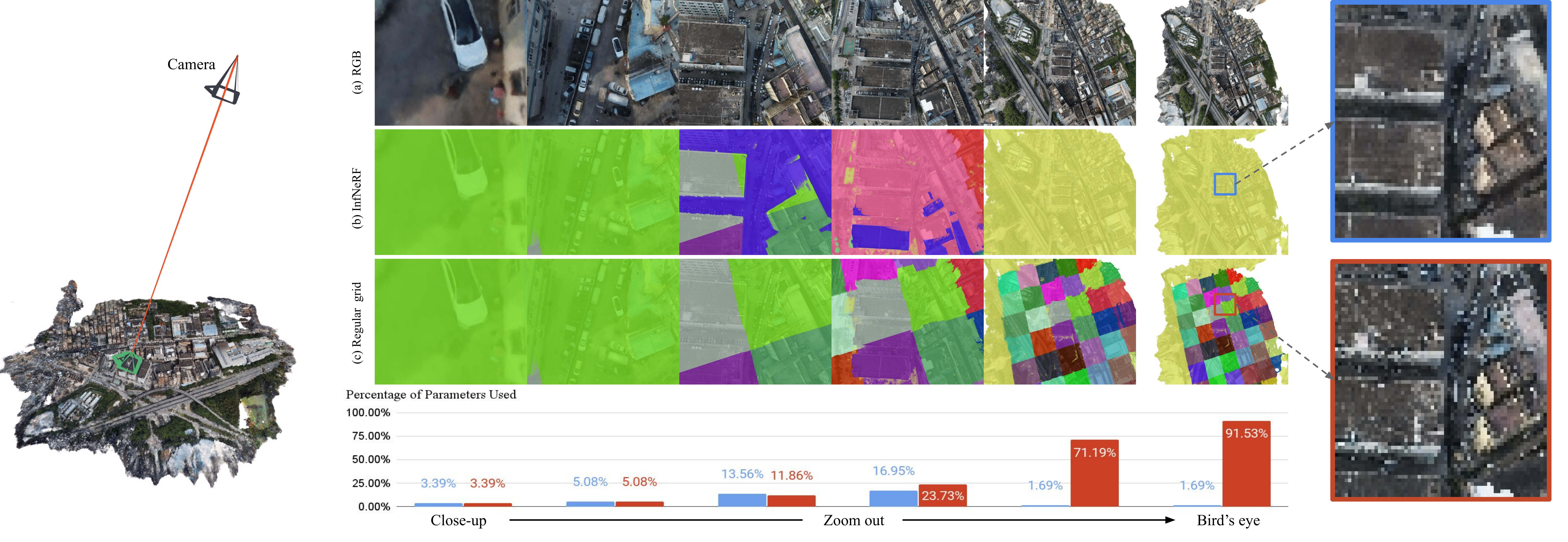}
  \caption{
  % With an LoD octree, InfNeRF divides the scene in space and scale into cubes. Each cube is individually represented by a NeRF.
  % This strategy enables the reconstruction of a large-scale scene with delicate detail.
  % During rendering, InfNeRF selectively fetches only a small portion of the model, significantly reducing memory footprint and accelerating data transfer time. 
  % InfNeRF's octree also enhances the anti-aliasing, resulting in superior image quality.
  % For example, a zoom-out trajectory from a close-up view to a bird's-eye view is shown on the left.
  % The first row in the middle shows the rendered RGB images from this trajectory. (a) shows InfNeRF's blocks by coloring each block with a different color while (b) shows a trivial grid partition's blocks.
  % Notably, only 16.95\% of blocks are required (bottom row, the 4th column), whereas the grid partition would require all blocks.
  % A part of the last frame is highlighted on the right column, with \textcolor{blue}{InfNeRF}'s anti-aliasing result at the top exhibiting less noise and a smoother appearance compared to the \textcolor{red}{grid partition}'s result shown at the bottom. 
    Efficient large-scale scene rendering with the proposed InfNeRF.
    (a) shows the rendered RGB frames by InfNeRF under a zoom-out camera trajectory. 
    Unlike existing methods that employ regular grid partition at a single scale (c), we propose a hierarchical NeRF framework to decompose the scene in both space and scale dimensions (b), in a spirit similar to the classical LoD techniques.
    This innovation is color-coded in (b) and (c), with each color representing a unique NeRF node within the whole model.
    During rendering, InfNeRF only requires a fraction of the nodes, considerably diminishing the memory footprint and expediting the model retrieval process.
    As demonstrated in the middle bottom chart, a mere maximum of 16.95\% parameters of the entire model are used for the rendering of InfNeRF, in stark contrast to the 91.53\% demanded by the baseline approach.
    In addition to the improved storage efficiency, the hierarchical structure of InfNeRF naturally improves its anti-aliasing capability.
    As shown in the rightmost column, InfNeRF shows notably reduced noise and enhanced smoothness while the existing methods suffer from severe aliasing artifacts.}
  \Description{figure description}
  \label{fig:teaser}
\end{teaserfigure}
\maketitle
\section{Introduction}
% \subsection{Large scale reconstruction}
% large scene reconstruction is important, useful.
% Recent advances in neural rendering techniques have lead to significant progress towards photo-realistic novel-view synthesis. 
% In particular, Neural Radiance Field(NeRF) has garnered significant attention, spawning a wide range of follow-up works that improve upon various aspects of the original methodology. 
% Its differentiable property makes NeRF well-suited for a diverse sorts of applications, such as 3D object generation. 
% Most importantly NeRF excels in generating more realistic images compared to traditional meshes, specially with transparent or reflective objects.
% However, most of NeRF studies focus on single object or small scene, the challenge of effectively representing large-scale scenes has not been thoroughly investigated or resolved.
Recent progress in Neural Radiance Fields (NeRF) has significantly advanced the field of photo-realistic novel-view synthesis~\cite {mildenhall2021nerf,barron2021mip,barron2023zip}. 
Surpassing traditional 3D representations like meshes, NeRF demonstrates a superior capability in generating accurate novel views, particularly for transparent or reflective geometries. 
However, despite these advancements, a predominant focus of NeRF research has been confined to single objects or limited-scale scenes. The challenge of effectively representing extensive, large-scale scenes remains largely underexplored and unresolved.

% Large scale scene reconstruction have wide-ranging applications, such as survey, search and rescue operations, construction planning, and navigation. 
% also under the name of novel-view synthesis is a practical topic, 
% One of the most famous applications is Google Earth, a platform that enables user traverse the Earth, to appreciate the magnificent beauty of nature and to explore the intricate human heritage.
Large-scale scene reconstruction has a wide range of applications, such as land surveys, search and rescue operations, construction planning, and navigation. A prime example of its potential can be seen in platforms like Google Earth, enabling users to virtually traverse the globe, which not only showcases the magnificent beauty of natural landscapes but also facilitates the exploration of intricate historic buildings.
Extending NeRF to handle large-scale scenes is intriguing and important because it opens up possibilities for more immersive and realistic representations of our world.

% \subsection{Rendering complexity}
% Trivially extending NeRF's capacity proves insufficient to address the challenge of rendering large-scale scenes, primarily due to the inherent space complexity. 
%As the scenes expand, it becomes impractical to store the entire model in memory of the rendering device such as mobile devices or consumer PCs.
% Consequently, a division of the model and storage on disk or in the cloud becomes necessary.
\subsection{Block-NeRF and Mega-NeRF}
Simply scaling up the architecture of NeRF proves insufficient to address this challenge, primarily due to the escalating space complexity.
As the scenes expand, it becomes impractical to store the entire model in the memory of standard devices such as mobile phones or consumer PCs.
Consequently, dividing the model and storing it in disk or cloud systems becomes necessary.

Towards this direction, recent large-scale NeRF approaches, such as Block-NeRF~\citep{tancik2022block} and Mega-NeRF~\cite{turki2022mega} divide the whole neural field equally in space into smaller, more manageable blocks, referred to as grid partition in this paper. 
This allows the rendering device to load only the data relevant to the camera's current location.
However, these methods encounter challenges in scenarios requiring a bird's eye view of the entire scene. 
Such a panoramic view plays a crucial role in human interactive navigation. For example, to adjust the position of the camera, it is much more efficient to first zoom out, identify the target location, and then zoom in again, compared to the laborious process of navigating at a constant zoomed-in level. 
In such scenarios, the existing methods need to load all blocks simultaneously, which can overwhelm the memory capacity of the device.
%todo rephase, jump to quick
In addition to the memory issue, these approaches are prone to aliasing artifacts that arise from an insufficient sampling rate for high-frequency signals as illustrated in Fig. \ref{fig:teaser}.

\subsection{Proposed InfNeRF}

%\begin{figure}[t]
%  \label{fig:octree}
%  \centering
%  \includegraphics[width=\linewidth]{image/octree.png}
%  \caption{Example of A two level octree, whose root represents the scene with a NeRF in a coarse manner. And each child represents a sub cube in a finer manner with a NeRF having similar amount of parameters. }
%  \Description{Example of multi-resolution octree}
%\end{figure}
\begin{figure}[t]
  \centering
  \includegraphics[width=0.8\linewidth]{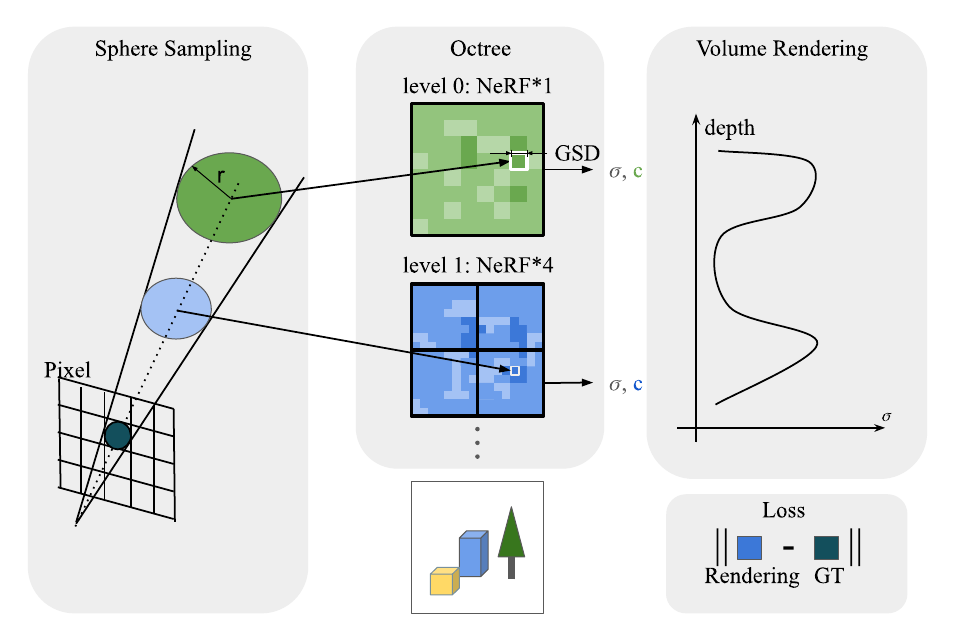}
  %\vspace{-1mm}
  \caption{Each sampling sphere along the ray is assigned a radius corresponding to its pixel size. The radius is proportional to depth. Depending on the radius, the sampling spheres will be sent to different nodes of the octree to query their densities and colors. The densities and colors are integrated using volume rendering to produce the final color for the pixel.   }
  \label{fig:arch}
  \Description{Example of multi-resolution octree}
  %\vspace{-4mm}
\end{figure}
\begin{figure}[h]
  \centering
  \includegraphics[width=0.8\linewidth]{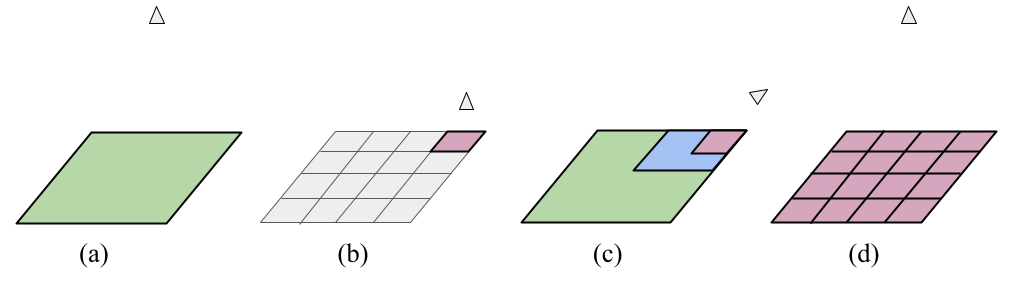}
  %\vspace{-1mm}
  \caption{As illustrated in (a), when zooming out, only the root node is required. In (b), when zooming in, only the leaf node is required. In (c), when looking at the horizon, approximately $\mathcal{O}(\log n)$ nodes are required which is the upper bound of InfNeRF. In (d), in contrast, other methods require all the blocks when zoom out resulting in an upper bound of $\mathcal{O}(n)$. }
  %\vspace{-4mm}
  \label{fig:lod}
  \Description{Illustration of LoD in InfNeRF}
\end{figure}

In this paper, we propose InfNeRF, a novel method that addresses the above challenges.
The core concept of InfNeRF is inspired by the classical mesh-based Level of Detail (LoD) techniques~\cite{clark1976hierarchical,erikson2001hlods, ribelles2010improved, tanner1998clipmap}. 
An overview is illustrated in Fig. \ref{fig:arch}.
Specifically, we construct a LoD tree for the target scene, where each node encapsulates a small NeRF with a roughly equivalent number of parameters.
The root node provides a coarse representation of the entire scene, while its successive children progressively describe more detailed sub-regions. 
During rendering, each sampling point is forwarded by a different NeRF, depending on its location and radius. 
The density and color then are accumulated to determine the pixel color.

As illustrated in Fig. \ref{fig:lod}, InfNeRF only requires a minimum subset of the nodes in rendering, chosen based on their proximity to the camera and their intersection within the frustum.
This significantly reduces the memory burden and the I/O time for retrieving parameters from disk or cloud storage. 
The memory complexity is notably efficient at $\mathcal{O}(\log n)$, where $n$ denotes the total information captured in the reconstruction.
In the real world, with $\log n \approx 20$, we can represent the Earth down to a resolution of centimeters.
In general, InfNeRF is like the combination of Mega-NeRF~\citep{turki2022mega} and PyNeRF. It divides the scene in both spatial and scale dimensions while existing approaches typically only consider one of them. 
%InfNeRF distinguishes itself from existing methods such as Block-NeRF~\citep{tancik2022block} and Mega-NeRF~\citep{turki2022mega} in that it subdivides the scene in both spatial and scale dimensions while existing approaches typically only consider the spatial aspect.

Another remarkable feature of InfNeRF is that the parent nodes in the LoD tree act as a smoother, low-pass-filtered version of the scene.
Therefore, the proposed method naturally removes high-frequency components in distant views, which effectively reduces the aliasing artifacts and enhances the visual quality of the generated images (Fig.~\ref{fig:teaser}).
In our experiments, InfNeRF achieves superior rendering quality for large-scene reconstruction, with an improvement of over 2.4 dB in PSNR while accessing only 17\% of the total parameters.

We also propose a pruning algorithm to prune the octree into an effective and compact representation using sparse points from SfM. 
This algorithm adapts the tree to each scene intelligently without any human intervention.

%We show that the total number of parameters of our model remains within $\mathcal{O}(n)$.
%In addition, to facilitate efficient training of InfNeRF, we introduce a novel training strategy that maintains a time and space complexity of $\mathcal{O}(n)$.
In addition, the training of InfNeRF only increases slightly and maintains a time and space complexity of $\mathcal{O}(n)$.
Moreover, this strategy allows for the training process to be effectively parallelized across machines. 

%InfNeRF distinguishes itself from existing methods such as Block-NeRF~\citep{tancik2022block} and Mega-NeRF~\citep{turki2022mega} in that it divides the scene in both spatial and scale dimensions while existing approaches typically only consider the spatial aspect. 

InfNeRF imposes no constraints on the underlying NeRF model and makes no assumptions about the scene, making it an adaptable framework that can seamlessly accommodate and leverage the rapid advancements in NeRF models.

In summary, our contributions are as follows:
\begin{itemize}
\item We propose a novel InfNeRF for large-scale scene reconstruction. By integrating neural fields with the classic LoD, InfNeRF achieves efficient rendering with a significantly reduced memory footprint.
\item Our InfNeRF demonstrates a substantial improvement in rendering quality, with over 2.4 dB increase in PSNR, and exhibits superior anti-aliasing capabilities.
\item We develop an efficient and scalable training strategy tailor-made for InfNeRF, addressing the complexities inherent in large-scale neural field training.
\end{itemize}

\section{Related Work}
\subsection{Photogrammetry}
Among the established reconstruction techniques, mesh-based reconstruction methods called photogrammetry have been used widely for it efficiency and robustness in creating digital twins world.
%This workflow includes successive steps: 
%Structure from Motion (SfM)~\citep{agarwal2010rome, wu2013vsfm, schonberger2016colmap,moulon2017openmvg} for camera pose estimation,
%Multi-View Stereo (MVS) techniques~\citep{hirschmuller2007SGM,bleyer2011patchmatch,schoenberger2016mvs} for depth computation, mesh reconstruction~\citep{kazhdan2006poisson, vu2011high}, 
%and texture mapping~\citep{waechter2014let}. 
%These steps form the basic workflow to create a textured mesh that can be efficiently rendered by a traditional graphics pipeline. 
Open-source tools, such as COLMAP~\cite{schonberger2016colmap} and OpenMVS~\cite{cernea2020openmvs}, have seamlessly integrated this workflow and gained widespread adoption in the academic community.
Furthermore, commercial software solutions like ContextCapture and DJI Terra~\cite {terra} have significantly elevated the robustness and efficiency to meet industrial standards.

\subsection{Level of Detail}
When the scene is large, it becomes impractical to store it entirely in memory.
In this case, the Level of Detail (LoD) or Hierarchical Level of Detail (HLoD)~\cite{clark1976hierarchical, erikson2001hlods,hoppe1998smooth,hoppe2023progressive,ulrich2002chunkedLoD,luebke2003level} provides a framework for efficiently storing, fetching, and rendering the scene.

Take tiled 2D map~\cite{wikitile} used widely in 2D navigation system for an example.
The maps in different resolutions are cut into 256$\times$256 PNG images referred to as tiles. 
These tiles are organized into a quadtree structure, typically comprising 22 levels. 
This hierarchical arrangement allows web browsers to rapidly fetch data and render any location at any zoom level with constant space complexity.

A similar technique can also be applied to 3D mesh.
After the reconstruction of the scene into one detailed mesh, 
this mesh undergoes multiple simplifications ~\cite{daniels2008quadrilateral} to reduce the number of triangles. 
Concurrently, its texture also undergoes downsampling multiple times~\cite{williams1983pyramidal}. 
Together this results in a mesh pyramid with varying levels of detail, ranging from fine to coarse.
Subsequently, these meshes are partitioned into numerous tiles and organized hierarchically, resembling a tree structure similar to a quadtree in 2D mapping.
%The physical size of the texel, referred to as GSD denotes the level of detail represented by this tile.
%Furthermore, multiple trees built from different image sources, such as drones, planes, or satellites, at various levels, can be combined into a larger tree to represent a larger scene. 
%, and finally form the earth.
During rendering, 3D engine such as  CesiumJS~\cite{web:cesium} or OpenSceneGraph~\cite{burns2004open} walks the tree, then progressively fetch tiles based on their proximity and intersection with the frustum.
This LoD technique enables users to have a seamless 3D navigation experience in a very large scene while maintaining a low memory footprint.
%if the tile is close and intersect with the frunstrum, the engin fetch it asynly.
%It is log n, to reduce the complexity, far plane clipping is apply, to refuse the tile too far. 
%But this lead to popup. some game engine will apply fog, to show the tile progressively and avoid hard popup. 
%and 3D mesh LoD like cesium. fluent web rendering.
\subsection{NeRF}
NeRF~\cite{mildenhall2021nerf} employs Multi-Layer Perceptrons (MLPs) for NeRF to capture a continuous density and viewpoint-dependent radiance of a scene. 
%Plenoxel introduces a voxel-dense grid~\cite{plenoxels}, which significantly improves the training and rendering speeds of NeRF, albeit with a significant VRAM requirement.
Methods like TensoRF~\cite{chen2022tensorf} and MERF~\cite{reiser2023merf} decompose the dense grid into low dimension feature tensor, while InstantNGP~\cite{muller2022instant} opts to store these vectors in a sparse hash table. These strategies greatly enhance model compactness while maintaining speed.

\subsection{Large-Scale Scene and Partition at Space Dimension}
While previous NeRF approaches primarily focus on scenes of limited scale, scaling up NeRF to handle large-scale scenes, such as cities or even the entire Earth, would empower a broader range of applications.
Mip-NeRF 360~\cite{barron2022mip} proposed a scene contraction technique to compress the unbounded scene into a bounded region.
However, this approach compromises quality in the compressed areas.
Grid-Guided NeRF~\cite{xu2023grid} extends its capacity by combining a 2D feature grid and MLP, but its reliance on a 2D plane assumption limits its capability to handling complicated 3D scene.
%Furthermore, these large-scale NeRF approaches still face challenges across different resolutions during the training and rendering phases.
Block-NeRF~\cite{tancik2022block}, Mega-NeRF~\cite{turki2022mega}, SMERF~\cite{duckworth2023smerf} and ScaNERF~\cite{wu2023scanerf} divide the scene into several sub-regions and reconstruct them separately. 
In its specialized indoor scenario, SMERF can even achieve impressive real-time rendering. 
However, all these methods suffer from a common issue: when rendering a bird's eye view, the entire model has to be loaded into VRAM, limiting their scalability.
Additionally, they may suffer from severe aliasing artifacts.

\subsection{Anti-aliasing and Partition at Scale Dimension} 
Anti-aliasing has been a fundamental problem in computer graphics and image processing for a long time and has been extensively explored in the rendering community. 
The common approaches to tackle this problem can be categorized into two ways: super-sampling and pre-filtering.
In the context of NeRF, super-sampling is casting multiple rays per pixel as in MobileNeRF~\cite{chen2023mobilenerf}, or sampling multiple points with perturbation, as demonstrated in Zip-NeRF~\cite{barron2023zip}.
This straightforward strategy is useful but also computationally expensive.
On the other hand, Mip-NeRF ~\cite{barron2021mip} introduces the concept of pre-filtering through the integrated position encoding.
Other research has explored scene partitioning at different scales to mitigate aliasing.
For instance, BungeeNeRF~\cite{xiangli2022bungeenerf} proposes a multi-scale MLP that progressively adds detail when the camera zooms in. 
%However, akin to Mip-NeRF, it may only accurately represent the center of the scene, lacking detail elsewhere.
Tri-MipRF~\cite{hu2023tri}, PyNeRF~\cite{turki2023pynerf} and Mip-VoG~\cite{hu2023multiscale} train a multi-resolution feature grid.
%Each of these methods showcases an impressive anti-aliasing.
These methods demonstrate impressive anti-aliasing capabilities and consistently high-quality rendering across the entire scene.
However, their lack of space partitioning leads to challenges when scaling up, requiring the whole finest layer to render a close-up frame.

In contrast to the methods mentioned in this subsection, the anti-aliasing problem is addressed with inherently no additional cost in InfNeRF.
Our parent node in the octree is a downsampled copy of the child nodes. 
Rendering these nodes at an appropriate zoom level yields a result free from aliasing artifacts.

\subsection{Scene Partition with an Octree}
Some researches, like PlenOctrees~\cite{yu2021plenoctrees} and NGLoD~\cite{takikawa2021nglod} utilize an octree-based approach to represent the scene. 
PlenOctrees stores the spherical harmonics coefficients in the leaves and uses the octree to accelerate accessing, while in NGLoD, a signed distance field is stored in the octree to represent a LoD closed volume. 
These direct storage techniques provide fast access and training but require significant memory, limiting their scalability. 
In contrast, InfNeRF employs many compact hash tables as the underlying data storage. The shallow octree is primarily used for organizing the hash tables into a LoD structure. This approach results in a much more compact LoD representation and much better scalability.

\subsection{Gaussian Splatting}
In recent times, Gaussian splatting~\cite{kerbl3Dgaussians} has gained significant attention due to its fast rendering speed compared to NeRF.
Additionally, the introduction of Mip-Splatting~\cite{yu2024mip}  effectively reduces aliasing artifacts across different resolutions.
Then Hierarchical Gaussians~\cite{hierarchicalgaussians24}, VastGaussian~\cite{lin2024vastgaussian}, and CityGaussian~\cite{liu2024citygaussian}, explore its potential in large scene reconstruction. While Gaussian splatting may not achieve the same PSNR as traditional NeRF~\cite{liu2024citygaussian} in challenging scenarios such as varying light conditions or camera exposure, it remains a promising technique for large-scale scene reconstruction in the future.
%Nowadays gaussian splatting get a lot of attention for its fast rendering comparing to NeRF. It has good anti-aliasing result ~\cite{yu2024mip} and can be applied to large scale reconstruction~\cite{hierarchicalgaussians24}. 
\section{Method}
%Recall that Google map, Google earth, Cesium can represent the whole earth with online streaming.
%we try to use an octree to achieve the same result which including
%total parameter O(n) possible for dynamice fetch.
%rendering in O(1)-O(log n) complexity. 
%The overview architecture of InfNeRF is illustrated in Fig. \ref{fig:arch}. InfNeRF divides the scene recursively into many cubes. Each cube is represented by a sub-NeRF. 
%The sampling point along the ray will be distributed to different NeRF, and accumulate their density and color to obtain the pixel color. in section \ref{subsec:pruning} will describe a way to prune the tree,  to achieve higher compactness. Then the tailored rendering algo is presented in \ref{subsec:render_pruned}. And finally we will present how we train the octree with an image pyramid.
% In Section \ref{subsec:tree}, we will present the octree structure, followed by an explanation of the anti-aliasing rendering process in Section \ref{subsec:rendering}.
% In Section \ref{subsec:pruning}, we detail the tree pruning methodology aimed at significantly enhancing sparsity and compactness. 
% In Section \ref{subsec:render_pruned}, we will discuss the necessary adaptations required for our rendering process to render the pruned tree.
% In Section \ref{subsec:treetraining} and \ref{subsec:dptraining}, we introduce our training approach, including the distributed training strategy.
% Finally, in section \ref{subsec:analysis}, we analyze the computational complexity of InfNeRF.

We present InfNeRF, a framework for efficient large-scale scene reconstruction using a LoD octree structure (Section~\ref{subsec:tree}). 
This tree structure naturally enables anti-aliasing rendering (Section~\ref{subsec:rendering}), which significantly enhances the visual quality of rendered images.
We also introduce tree pruning to improve model sparsity and compactness, along with a rendering approach tailored for the pruned structure (Section~\ref{subsec:pruning}). 
Moreover, we propose an efficient training strategy for InfNeRF in Section~\ref{subsec:treetraining}. 
Finally, we provide a comprehensive analysis of the computational complexity of InfNeRF in supplementary material, offering insights into the efficiency and scalability of our approach in handling large-scale scene reconstructions.

\subsection{Tree Structure}\label{subsec:tree}
The core of our InfNeRF is a LoD structure, essentially an octree where each node represents a specific cubic space of the scene, known as the node's Axis-Aligned Bounding Box (AABB). The octree is illustrated in the middle of Fig. \ref{fig:arch}.
InfNeRF begins with the entire scene encapsulated within a root node, which is then divided into eight smaller cubes, corresponding to the child nodes.
This division process continues recursively, partitioning the scene into increasingly finer parts, both in terms of space (size of the area) and scale (level of detail).
%The octree is composed of numerous nodes, each corresponding to a cubic space of the scene, referred to as the node's Axis-Aligned Bounding Box (AABB). 
%This cubic AABB is equally divided into 8 sub-cubes, corresponding to the node's 8 child nodes. 
%Recursively, the scene is divided in both space and scale into cubes.

% Each node in the tree then reconstructs its AABB with its NeRF independently.
% We do not impose any requirement on the NeRF methods. 
% It can take various forms, such as an MLP~\citep{mildenhall2021nerf}, Instant-NGP~\citep{muller2022instant}, tri-plane~\citep{Chan2022,hu2023tri}, or even a combination of these methods, as long as it can query the density $\sigma$ and the color $\mathbf{c}$, given a 3D point $\mathbf{x}$ and direction $\mathbf{d}$ as:
%who can be a MLP, Instant NGP, tri-plane or even mixed, only if it can query the density and RGB giving a 3D point and direction.

Each node in this tree is paired with its own NeRF, which represents the node's AABB at a certain scale. 
Our approach is flexible regarding the choice of NeRF model. 
It can be a simple MLP~\citep{mildenhall2021nerf}, Instant-NGP~\citep{muller2022instant}, triplane~\citep{Chan2022,hu2023tri}, or even a combination of them, as long as it can query the density $\sigma$ and the color $\mathbf{c}$ for a given 3D point $\mathbf{x}$ and viewing direction $\mathbf{d}$:
\begin{equation}
    \sigma, \mathbf{c} = \text{NeRF}(\mathbf{x}, \mathbf{d}).
\end{equation}

In this work, we use Instant-NGP for its efficiency in training.
Instant-NGP discretizes each cube into a grid of voxels, and the number of voxels along each axis of the cube is defined as grid size.
Correspondingly, the Ground Sampling Distance (GSD) of a node can be approximated as: 
\begin{equation}\label{equ:GSD}
 \text{GSD} = \frac{\text{length of AABB}}{\text{grid size}},
\end{equation}
whose unit is in meters.
The GSD, in this context, serves as a measure of resolution or granularity, indicating the level of detail that the node can capture about the physical world - the smaller the GSD, the finer the details. 
For example, if a node $\mathcal{A}$ represents a 1 km$^3$ cube with a grid size of 2048, the GSD will be $\frac{1000}{2048}\approx0.48$ meters.
As each node in our octree shares the same grid size, the child nodes of $\mathcal{A}$ have a GSD of $0.24$ meters, thereby capturing more detailed information.
This is also illustrated in Fig.~\ref{fig:arch}, where the GSD of a cube at level 1 is half that of a cube at level 0.

\subsection{Anti-Aliasing Rendering}
\label{subsec:rendering}
% In the original NeRF paper, the pixel RGB is obtained by accumulating the density and color of the sampling points along the ray.
% Our rendering is similar, with the key difference that the sampling points are queried from different nodes of the tree rather than a single NeRF as shown in Fig.~\ref{fig:arch}.
Similar to the original NeRF, the rendering process of InfNeRF accumulates the density and color of sampling points along a ray to determine the RGB of a pixel.
However, the key difference is that the sampling points are queried from different nodes of our octree as shown in Fig.~\ref{fig:arch} rather than a single NeRF .
In the following, we show how we find the corresponding node for a given sampling point.

According to the sampling theory~\citep{williams1983pyramidal}, a sample on the ray is not merely an infinitely small point, but rather a Gaussian sphere with a certain volume. 
For simplicity, we model this as a sphere with radius $r$ that corresponds to the pixel size,
\begin{equation}
  \label{equ:radius}
    % r = \text{depth} \times \frac{\text{pixel width}}{\text{focal length}},
    r \approx \text{depth}\times \frac{\text{pixel width}}{2\times\text{focal length}},
\end{equation}
as illustrated on the left of Fig. \ref{fig:arch}.
Then, given a sampling sphere with position $\mathbf{x} \in \mathbb{R}^3$ and radius $r \in \mathbb{R}$, we aim to find the node whose granularity (GSD) matches $r$ and whose AABB contains $\mathbf{x}$. 
In this way, we ensure InfNeRF samples a proper region in the 3D space rather than just a single point, leading to a desirable anti-aliasing effect.
% In this case, the sampling region precisely aligns with a single voxel stored in the NeRF, leading to a perfect anti-aliasing effect.

% However, the GSD is discrete in our tree structure. 
% We can either sample from two levels and interpolate it, as PyNeRF~\cite{turki2023pynerf} did, or we can round it to one level.
% InfNeRF rounds it toward the root for computational efficiency.
%In our observation, we have noticed that when the sampling sphere falls between two discrete levels, it can lead to small aliasing artifacts. Additionally, when sampling points transition between levels, discrepancies in density or color may arise, resulting in an undesirable "popup" effect when the camera moves.
We observe that when the sampling sphere falls between two discrete levels, small aliasing occurs. Nevertheless when sampling points transition between levels, discrepancies in density or color may arise, engendering an undesirable "popup" effect when camera moves. 
To mitigate these phenomenons, both PyNeRF and Tri-MipRF employ interpolation techniques between adjacent levels which will double the computation.
However, we pursue a simpler approach by introducing stochastic perturbations to the radius of the sampling sphere.
\begin{equation}
    r_{prt}=r \times 2^p, \text{ where }  p \sim \mathcal{U}(-0.5,0.5).
\end{equation}
This strategy use human eye's natural blending effect to attenuate the abrupt transitions without incurring additional computational cost.
Finally, the level to sample a given sphere is determined by:
% So the level to forward the sphere can be found by:
\begin{equation}
  \label{equ:level}
       \text{level} = \lfloor \log_{2}{\frac{root.gsd}{r_{prt}}} \rfloor,
\end{equation}
where $root.gsd$ denotes the GSD of the root node, \textit{i.e.}, level 0.

% Here, level 0 corresponds to the root of the tree.
As illustrated in Fig.~\ref{fig:lod}(a), when the camera is far away from the scene, all spheres are relatively large, and therefore, their densities and colors are estimated by the root node, resulting in a smooth anti-aliasing effect. 
Conversely, as shown in Fig.~\ref{fig:lod}(b), zooming in with the camera reduces the size of the spheres, which shifts the estimation to the leaf nodes, yielding a sharp and detailed image.
As InfNeRF necessitates only a minimum subset of the nodes for rendering a frame, it requires loading merely $\mathcal{O}(\log n)$ number of parameters from the whole model into memory.
The upper bound $\mathcal{O}(\log n)$ is reached when the camera looks at the distant horizon, as illustrated in Fig.~\ref{fig:lod}(c).
In this scenario, multiple-level nodes are engaged simultaneously, from the root node representing the blurring horizon to the leaf node representing foreground details.
% Similarly, in Fig.~\ref{fig:lod}(b), when a user zooms in to explore specific details, bringing the scene closer, the spheres become smaller and are directed to the leaf nodes. 
% This yields a detailed and sharp image.
%the root node representing the blurring horizon, the leaf node representing foreground detail, and all the intermediate nodes in between, are required at the same time.

\subsection{Tree Pruning}\label{subsec:pruning}
%\begin{algorithm}\label{algo:recursive build}
%\caption{recursive octree build}
%\begin{algorithmic}
%    \Function{Build}{s, AABB}\Comment{xxx}
%        \If{$\left|s\right|$ < thredshold}
%            \Return None
%        \EndIf
%        \State $node \gets new NeRF(AABB)$
%        \State $s \gets s - \{(x,r)|(x,r)\in s \text{ and } r> node.gsd\}$
%        \For{$i \gets 1$ to $8$}                    
%            \State $ child\_s \gets\ \{(x,r)| x \in s \text{ and } x \in subcube(AABB, i)\}$
%            \State $node.children[i] \gets \Call{Build}{child\_s, subcube(AABB,i)}$
%        \EndFor
        
%        \Return node
 %   \EndFunction
%\end{algorithmic}
%\end{algorithm}
While we can build a perfect octree as in Section~\ref{subsec:tree}, this may not be efficient, particularly for non-uniformly sampled scenes.
To optimize the structure, it is important to identify the essential nodes necessary for rendering all input images, effectively pruning the superfluous parts of the tree.

In this work, we utilize sparse points $ \mathbf{x} \in \mathbb{R}^3$ obtained from the Structure from Motion (SfM) to approximate scene geometry.
Each sparse point $\mathbf{x}_i$ observed in $M_i$ images is treated as $M_i$ spheres, each with a radius $r_{ij}, j=1,\cdots, M_i$ as defined in Section~\ref{subsec:rendering}.
We find the corresponding nodes for all the $\sum M_i$ spheres in an ideal, infinitely deep octree as described in Section~\ref{subsec:rendering}.
We retain only the nodes that correspond to sparse spheres and their respective parent nodes, while the others are pruned away.
% In this study, we utilize sparse points $ \mathbf{x} \in \mathbb{R}^3$, derived from the Structure from Motion (SfM) technique, as a means to approximate scene information.
%First We define the foreground AABB by 5\% and 95\% percentile of x in each axis.
%Each sparse point, observed by certain cameras, is also treated as a sphere $s = \{(x,r)| x \in\mathbb{R}^3, r \in \mathbb{R}\}$ with a radius $r$ corresponding to the size of the pixel, similar to the sample sphere in section \ref{subsec:rendering}.
%Each sparse point $\mathbf{x_i}$, observed by $M_i$ images, is treated as $M_i$ sphere with a radius $r$ defined in section \ref{subsec:rendering}. 
%Then imagine rendering these $\sum M_i$ spheres through an infinitely deep perfect octree in the way mentioned in section \ref{subsec:rendering}. Only the nodes reached by these sparse spheres are required to be constructed, along with their parent nodes. The rest of the nodes are pruned.

Compared to Mega-NeRF or Block-NeRF, which split the scene uniformly based on space occupancy, our approach offers a more intelligent resource allocation strategy. 
It adaptively creates deeper branches in areas with rich details and shallower branches in coarse regions. 
%In addition, our solution provides users with the flexibility to control the depth of the tree, enabling a tailored balance between reconstruction quality and computational efficiency.
%The foreground AABB then is defined by the 5\% and 95\% percentiles of $x$ in each axis.
%We define the foreground AABB by 5\% and 95\% percentile of the sparse points in each axis.
%The root node is created with a cubic AABB containing the entire foreground box. 
%These sparse spheres $s$ are sent to root. 
%For each node, if number of sparse spheres $\left|s\right|$ it receives surpasses a predefined threshold $th$, indicating there is enough visual information, the node therefor is created.
%The sparse spheres whose radius is smaller than GSD are split into 8 and recursively sent to 8 children.
%The recursive construction will eventually stop, even without user defined max depth, because the minimal radius is larger than 0 and $\lim GSD \rightarrow 0$. 
%Figure \ref{fig:pruninging} illustrates this process with a quadtree.
% Compared to MegaNeRF or BlockNeRF, which split the scene equally based on space occupancy, our solution offers a more intelligent resource allocation strategy. 
% The algorithm creates a deeper branch in regions with more details and a shallower branch in surrounding areas. 
% Users also have the flexibility to limit the tree depth based on the desired reconstruction quality.

This approach is robust as the subtree pruning is only restricted to cases where there are no sparse points in a relatively large cube.
As further detailed below, even in the rare case that a subtree is pruned by mistake, like in low texture area, the affected volume can still be well reconstructed by its parent node, albeit in a coarser manner. 
This ensures that, while fine details may be omitted, the essential structure is preserved. 

%This approach is robust as the subtree pruning only occurs when there is no sparse point in a relatively large cube.
%\xy{what is "2d feature triangulated"? this just comes from nowhere}
%This approach is robust as subtree pruning occurs only when very little sparse spheres exist in a $\text{grid size}^3$ volume, which also means this volume has less then $\text{th}$ feature point in a $\text{grid size}^2$ area of all observed image. 
%It is rare in practice where $\text{grid size}$ is often larger then 512. 
%The proposed process is robust, cause only if none sparse sphere existed in $grid\_size^3$, this subtree will be pruned, which is very rare when $grid\_size$ is large. 
% As further detailed below, even if a subtree is pruned by mistake, the corresponding volume can still be reconstructed by its parent, albeit in a coarser manner. 
% This ensures that while fine details may be omitted, the essential structure of the object is retained. 
%Because most of the sparse points lay on the surface of the object, which is a 2-manifold, so theoretically the octree will converge to a quadtree when $grid size \rightarrow 0$.
%It is also observed in experiment.

% add and image on garden

% \subsection{Pruned Tree Rendering}\label{subsec:render_pruned}
%Rendering a perfect octree is straightforward, each sample point can be trivially find the node to forward; however, the tree's efficiency is compromised due to significant capacity wasted in void or blur spaces. 
% \paragraph{Pruned Tree Rendering}
% \vspace{1mm}
% \noindent \textbf{Pruned tree rendering.} 
\subsubsection{Pruned Tree Rendering}
The pruned tree structure introduces new challenges in InfNeRF rendering, as the optimal node determined by Eq.~\ref{equ:level} may no longer exist after pruning.
In such scenarios, we instead process the sampling sphere with the nearest available ancestor of the intended node.

Specifically, given a sampling sphere $(\mathbf{x}, r)$, the rendering process recursively travels down the tree.
If the sphere successfully arrives at the node of the expected level, the density and color of the sphere will be estimated by that node. 
However, if this process is interrupted by a pruned node, the rendering of the sphere reverts to its parent node. 
We outline this rendering strategy with the pseudo-code in Algorithm~\ref{algo:recursive forward}.
%Finally, each sampling sphere is processed by the NeRF of the located node, ensuring accurate and efficient rendering despite the pruned tree structure.
The additional computational overhead introduced by tree traversal includes $log(n)$ times floating-point comparisons and memory reads, which can be considered negligible when compared to the NeRF evaluation.

\RestyleAlgo{ruled}
\SetKwComment{Comment}{//}{}
\begin{algorithm}
% \caption{recursive octree forward}
  \caption{Forward run of the pruned tree in Section~\ref{subsec:pruning}, starting from $node=root$.}
  \label{algo:recursive forward}
  \SetAlgoLined
  \KwIn{node}
  \KwIn{\textbf{x}, r, \textbf{d} $\gets position, radius, direction$}
  \KwOut{density, color}
  \SetKwFunction{KwForward}{Forward}
  \SetKwProg{Function}{Function}{:}{}
  \Function{\KwForward{node, \textbf{x}, r, \textbf{d}}}{
    \eIf{$r \geq node.gsd$}{
        \Return $node$.NeRF(\textbf{x}, \textbf{d}) \Comment*[r]{Process}
    }
    {
        $i \gets FindSubcube(node.AABB, x)$\;
        $child \gets node.children[i]$\;
        \eIf{$child$ = $nil$}{
            \Return $node$.NeRF(\textbf{x}, \textbf{d}) \Comment*[r]{Revert}
        }{%else
            \Return{\KwForward{$child$, \textbf{x}, r, \textbf{d}}} \Comment*[r]{Descend}
        }
    }
  }
      
\end{algorithm}
%\vspace{-5mm}

%The only different is that now the density and color are from different nodes of the tree not a single network.
%Since in InfNeRF each sample go through only one NeRF, the gradient of a pixel also goes to one NeRF.
%
\subsection{Training}\label{subsec:treetraining}
\subsubsection{Training Loss}
To train the proposed InfNeRF, we accumulate the density and color of all spheres sampled on a ray using volume rendering~\citep{mildenhall2021nerf} and compare the resulting RGB values with the pixels of the observed images using L2 loss $\mathcal{L}_{\text{RGB}}$. 
To remove the cloudy effect from the unobserved areas, particularly in the sky, and enhance visual quality in the bird's-eye view, we introduce a transparency loss $\mathcal{L}_\text{trans}=-\|\exp(-\sigma(\mathbf{x}))\|_1$, which uniformly samples points within the AABB and uses L1 loss to encourage their density to approach 0.
% \begin{equation}
%     \label{equ:trans}
%     \mathcal{L}_\text{trans}=-\|\exp(-\sigma(\mathbf{x}))\|_1.
% \end{equation}

To further improve the result, we also incorporate the distortion loss $\mathcal{L}_{\text{distort}}$ from~\citep{barron2022mip} and a regularization loss $\mathcal{L}_{\text{reg}}$ who measures the L2 differences of the density between levels, to promote the density consistency.% across different scales by encouraging the same locations from different scales have similar density.
The total loss of our training is summarized as:
\begin{equation}
    \label{equ:totalloss}
    \mathcal{L}_{\text{total}}=\mathcal{L}_{\text{RGB}} + w_{1}*\mathcal{L}_{\text{distort}} + w_{2}*\mathcal{L}_{\text{reg}} + w_{3}*\mathcal{L}_{\text{trans}}
\end{equation}

\subsubsection{Pyramid Supervision.} 
The images in our datasets are captured by drones at relatively close proximity, resulting in a lack of multi-resolution information necessary to supervise the higher-level nodes.
%While this offers sharp and detailed pixels that can well supervise the lower-level nodes, it leads to insufficient training for the higher-level nodes close to the root, which require low-frequency supervision signals.
To mitigate this issue, following PyNeRF~\cite{turki2023pynerf}'s approach,  we augment the training set using an image pyramid constructed by repeatedly downsampling the high-resolution images. 
The lower-resolution image with a larger pixel size, create larger sampling spheres according to Eq.~ \ref{equ:radius}. 
These larger sampling spheres are sampled from higher-level nodes in the octree, effectively supervising their training.
%Given the hierarchical nature of our tree structure, which has more lower-level nodes than the higher-level ones,
%the pixel sampling strategy of conventional NeRF, which samples all images equally, is not suitable for our pyramid supervision.
%
To account for the pixel imbalance across different pyramid levels, we perform uniform sampling in pixel domain, where a high-resolution image receives four times the sampling rate compared to its low-resolution counterpart. 
This strategy facilitates more balanced training across different levels of the image pyramid.

\subsubsection{Distributed Training}
As the scene grows, the demands on VRAM and training time also increases. 
Straightforwardly using existing distributed training approaches, such as Distributed Data Parallel (DDP), leads to increased VRAM consumption and substantial communication overhead.
To address these limitations, we propose a new distributed training strategy.

\begin{figure}[t]
  \centering
  \includegraphics[width=0.9\linewidth]{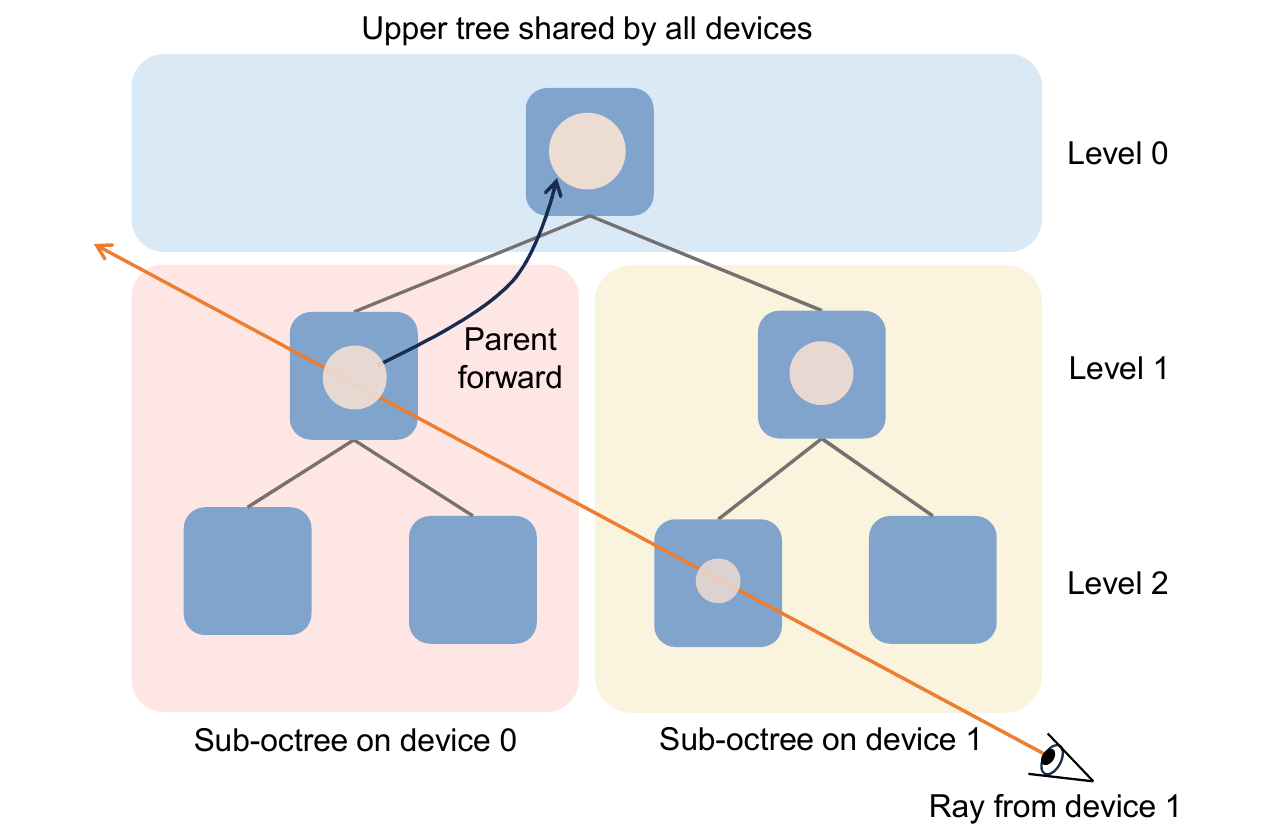}
  %\vspace{-1mm}
  \caption{Illustration of the distributed octree as a binary tree for simplicity. The tree can be divided into the root and 2 branches. The root is shared by all devices like Conventional DDP. And each branch is owned by one device without sharing the weights. When a sampling sphere descends to a pruned node, its density and color will be estimated by the shared parent node.}
  %\vspace{-4mm}
  \label{fig:hardcut}
  \Description{Illustration of the distributed training in InfNeRF}
\end{figure}

Given a specific tree level $L \in \mathbb{Z}^+$, we split the octree into two parts: the upper tree which includes all nodes from level 0 to level $L-1$, and the lower part which is a forest including a maximum of $8^L$ subtrees. 
%Evidently the AABBs of these $8^L$ subtrees form a partition of the scene's AABB in space.
During training, the upper tree is shared across all devices, and the parameters are updated concurrently. 
Meanwhile, individual subtrees
%or clusters of adjacent subtrees in the lower part,
are assigned among different devices and updated independently.
In other words, each device only maintains the shared upper tree and a subset of the lower subtrees, effectively reducing the VRAM footprint and communication overhead. 
For instance, let's consider a perfect octree with four levels. When the root node is shared $(L=1)$, the shared upper tree contains only a tiny fraction, approximately 0.17\%, of all the parameters in the entire tree.
%Take a 4 levels perfect octree for example. when $L=1$, the shared root node only process $ 0.17\% $ of all parameters.
%Each device own the upper tree, and his own subtree and responsible the train this space with it's own subtree and the shared upper tree.
% 
During forward pass, if a sampling sphere descends to a subtree that is not owned by the current device, it will be forwarded by its parent node in the shared tree as demonstrated in Fig.~\ref{fig:hardcut}.
%So in the forward part, there is will not generate overfitting noise in the foregournd.
Consequently, during training, the only data transfer is the all-reduce operation on the gradient of the shared upper tree, which leads to a more efficient and scalable training process.
% This configuration reduces the VRAM consumption on each device and communication overhead across devices while keeping global consistency. 
%We observed that sharing only the first level of the octree (set $L = 1$)  will largely improve the efficiency while keeping good consistency.

%We've also made the size of root and branches adjustable to fit different hardware capabilities.

Furthermore, the training data can also be divided among devices for further acceleration. 
Intuitively, if a ray on a particular device does not intersect with the AABB of any subtree on that device, it would be not sensible to sample the corresponding pixel of the ray.
Therefore, we find a 2D AABB for each device by projecting the sparse points inside the subtrees' 3D AABBs back onto images.
Then, during the training phase, only the pixels that fall within the 2D AABB are sampled and utilized. The experimental result of distribution training is provided in supplementary materials. 
\section{Experiment}
InfNeRF is implemented using the NeRFStudio framework~\cite{nerfstudio}. 
Each node's NeRF is an Instant-NGP with default parameters such as a grid size of 2048 and 16 levels of hash tables, each with a size of $2^{19}$.
Following NeRF-w~\cite{martinbrualla2020nerfw}, each image is assigned an appearance embedding vector $e \in \mathcal{R}^{32}$ to encode its lighting condition.
%The tree is limited to 4 levels, resulting a reasonable resolution of 5cm for a scene spanning 1 km.
More implementation details such as learning rate and experiment settings are provided in supplementary materials.

\subsection{Experiment On Large-Scale Dataset}
\subsubsection{Dataset}
To evaluate the scalability of our method, we adopt a similar approach to Mega-NeRF \cite{turki2022mega}, by conducting our main experiment using four real-world large-scale urban scenes dataset captured by drone from Mill 19\cite{turki2022mega} and UrbanScene3D \cite{UrbanScene3D}. 
Each dataset consists of around two thousand 4K images, amounting to 40 gigapixels and spanning 1 km. 
Notably, they are approximately 200 times larger than the scenes reported in other general NeRF work. 
Our octree is limited to 4 levels, resulting a reasonable GSD ranges from 0.4 m of the root node to 5 cm of the leaf node. The total model size is about 3 GB.
%Additional information about the dataset can be found in the supplementary and our webpage.
%The camera poses and the sparse points for our dataset were obtained using COLMAP \citep{schonberger2016colmap}. 
%Additional experiments and results on general scene datasets, and techniques for further improvements can be found in the supplementary and our webpage.

%The tree is limited to 4 levels, resulting a reasonable resolution of 5cm for a scene spanning 1 km.
%The source code and trained models will be released to the public.
%\subsection{Implementation}\label{subsec:implement}
%Inf-NeRF is implemented using the NeRFStudio framework~\citep{nerfstudio}. 
%Each node's NeRF is an Instant-NGP implemented by tiny-cuda-nn, with default parameters such as a grid size of 2048 and 16 levels of hash tables, each with a size of $2^{19}$.
%This configuration is reported to yield a similar PSNR as the MLP with comparable capacity, as mentioned in \citep{muller2022instant}.
% single appearance embedding is shared by the entire tree.
%The images from the same pyramid with different resolutions share one appearance embedding feature.
%The tree is limited to 4 levels (depth = 3), and a 5-level image pyramid is used in training, starting from the original resolution.
%Loss weights are set to $w_1=0.002, w_2=0.01, w_3=0.001$.

\subsubsection{Result on Space Complexity}
\begin{figure}
  \centering
    \includegraphics[width=0.8\linewidth]{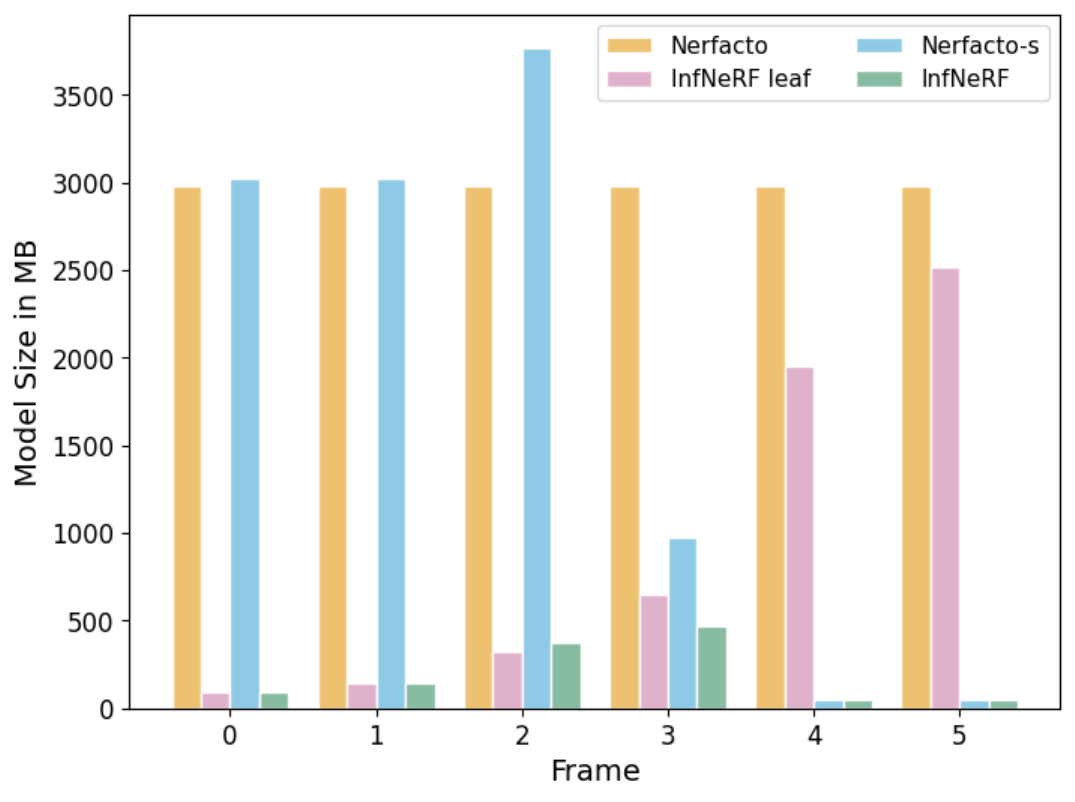}
  %\vspace{-1mm}
  \caption{We render a video starting from a close-up view and zooming out to the full scene and report the size of all sub NeRFs required for rendering each frame.
  Thanks to the LoD tree structure, InfNeRF demonstrated a much smaller memory footprint.} 
  %\vspace{-4mm}
  \label{fig:memory}
  %\Description{}
\end{figure}
%The paramount objective of our study is to improve the scalability of NeRF in Rendering. 
The paramount objective of our study is to reduce the required model size in rendering, which refers to the sum of all sub NeRFs' size required for rendering one frame. It is the total information that needs to be retrieved from the clouds and stored in the local device during real-time rendering.
%It determines how much information must be retrieved through the network and stored in the GPU. 
%So we compare InfNeRF to three other methods with similar model sizes, but different structure to demonstrate how the structure affect the rendering memory footprint. The three methods including 
To demonstrate the impact of our octree on the required model size, we compare InfNeRF with three other methods with similar total model sizes but with different architectures. These three methods are:
\begin{enumerate*}
    \item Nerfacto~\cite{nerfstudio},
    \item Nerfacto-s, a 4 levels Nerfacto, representing the scale partition methods like PyNeRF \cite{turki2023pynerf},
    %\item PyNeRF, our implemented 
    %\item InfNeRF,    
    \item InfNeRF leaf, an InfNeRF without upper tree structure, representing the space partition methods like Mega-NeRF~\cite{turki2022mega}.
\end{enumerate*}

In this experiment, we examine the rendering of a zoom-out trajectory, starting from a close-up view and transitioning to a bird's-eye view in VGA resolution. 
The trajectory and the corresponding rendered images are shown in Fig. \ref{fig:teaser}.
%, and a video demonstration is provided in the supplementary material.
%In this experiment, a zoom-out trajectory from a close-up view to a bird's-eye view in VGA resolution is rendered by these 4 methods. 
%The trajectory and the rendered images are shown in Fig. \ref{fig:teaser} and in the video of supplement material.
%To evaluate the space complexity of the different methods, we present the required model size in megabytes for six keyframes in Fig.~\ref{fig:memory}.
We present the required model size in megabytes for six keyframes in Fig.~\ref{fig:memory}.
%The model size in Megabytes required for 6 keyframes is presented in Fig \ref{fig:memory}.
When rendering the close-up view, the space partition method InfNeRF leaf proves advantageous as it only requires a single node of the scene and has low space requirements. 
In contrast, the scale partition methods Nerfacto-s necessitate the detailed version of the entire scene, resulting in a higher space complexity.
As the camera moves away, the space partition method InfNeRF leaf now requires all blocks to render the complete scene, while the scale partition method reduces its requirement by only rendering the low-resolution node.
%For rendering the close-up view, the space partition method require only one part of the scene and has low space reqirement. 
%Contrastly the scale partition methods require the detailed version of the whole scene, leading to a high space requirement.
%As the camera move away, The space partition method require all block to render the whole scene, and the scale partition method require only the low resolution portion  
However, InfNeRF consistently exhibits its low space complexity in both scenarios, with a memory footprint of only 17\% compared to InfNeRF leaf and 12.3\% compared to Nerfacto-s.
%InfNeRF exhibits its low space complexity in both case, it has a memory footprint that is only 17\% of InfNeRF leaf and 12.3\% of Nerfacto-s.

\subsubsection{Result on Rendering Quality}
%Quantitative results are reported based on metrics such as PSNR, SSIM, and LPIPS using the Alex implementation.
As zoom-out is a crucial use case in our paper, we not only report metrics at the finest resolution, denoted as PSNR0, but more importantly, the mean metrics over 6 resolution levels, denoted as PSNR, SSIM, LPIPS.
%For training the appearance embedding, similar to MipNeRF~\cite{barron2021mip} and Mega-NeRF~\cite{turki2022mega}, the left half of the test image is used in training.
%Evaluation is conducted solely on the right half of the image. 
We compare our approach with
\begin {enumerate*}

\item Nerfacto~\cite{nerfstudio}, whose model size is scaled up to InfNeRF,
\item PyNeRF~\cite{turki2023pynerf}, also being scaled up to InfNeRF,
\item Mega-NeRF~\cite{turki2022mega} with 25 partitions,
\item InfNeRF compact (InfNeRF-c), whose model size is similar to Mega-NeRF.
\end {enumerate*}
%As this large model takes almost one month to train with 1 GPU, we evaluate the largest and best checkpoint available on their website.
%In addition to these two methods, we also train and evaluate a smaller version of InfNeRF, called InfNeRF compact (InfNeRF-C), whose model size is similar to Mega-NeRF.

%Warameters.hen render Infnerf use same amount of parameter as Mega-NeRF

%Our leaf is same parameter of our full, but disable all the non leaf node. To compare anti alias result.

\begin{figure}
    \centering
    \includegraphics[width=0.7\linewidth]{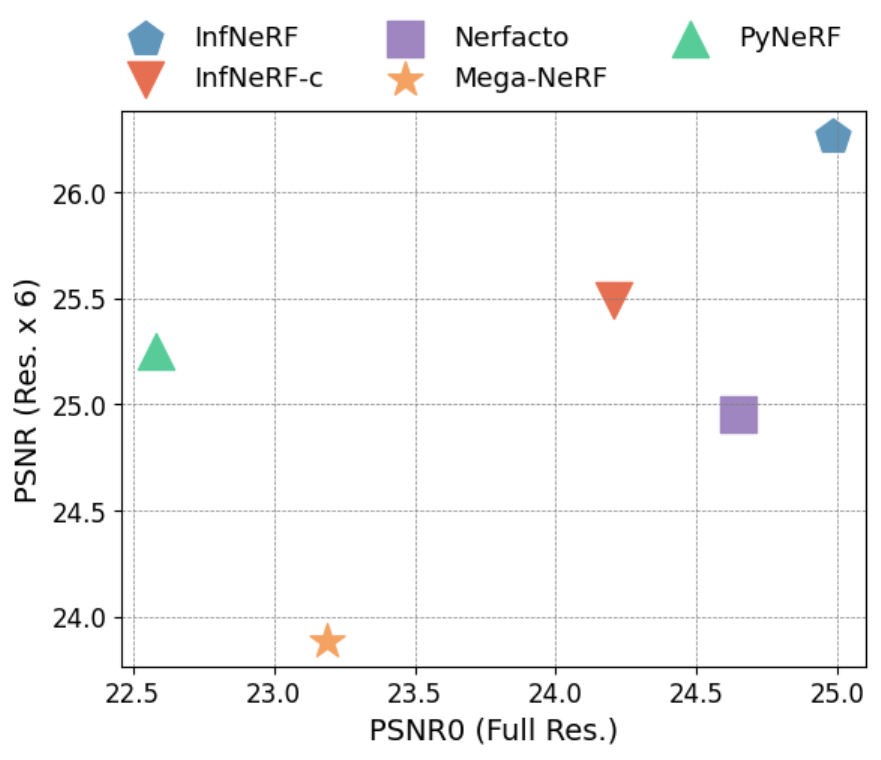}
    %\vspace{-3mm}
    \caption{The PSNR for full resolution and across 6 resolutions of 4 methods are presented in the above plots. InfNeRF is superior to other methods in both dimensions.}
    %\vspace{-3mm}
    \label{fig:2psnr}
\end{figure}

%\begin{table*}
%  \caption{Quantitative comparison on four public large-scale scene datasets.  We report PSNR0 from full resolution and mean PSNR, SSIM, and LPIPS metrics from 6 resolutions. The \textbf{best results} are highlighted.}
%  \label{tab:comparison}
%  \begin{tabular}{c|cccc|cccc}
%    \toprule
%    Scene&\multicolumn{4}{c|}{Residential}&\multicolumn{4}{c}{Sci-Art}\\
%    Metric&PSNR0$\uparrow$&PSNR$\uparrow$&SSIM$\uparrow$&LPIPS$\downarrow$&PSNR0$\uparrow$&PSNR$\uparrow$&SSIM$\uparrow$&LPIPS$\downarrow$\\
%    \midrule
%Nerfacto     &24.32  &24.31  &0.772  &0.216  &26.75  &26.48  &0.827  &\textbf{0.182}\\
%    Mega-NeRF       &22.52  &22.85  &0.748  &0.265  &24.37  &24.22  &0.808  &0.202\\
%    InfNeRF-c    &23.84  &24.87  &0.785  &0.198  &25.83  &25.81  &0.805  &0.212\\
%    %Ours leaves \\
%    InfNeRF   &\textbf{24.66}  &\textbf{25.46}  &\textbf{0.820}  &\textbf{0.158}  &\textbf{27.03}&\textbf{26.96}&\textbf{0.839}&0.1918\\
%    \midrule
%    Scene&\multicolumn{4}{c|}{Rubble}&\multicolumn{4}{c}{Building}\\
     %Metric&PSNR0$\uparrow$&PSNR$\uparrow$&SSIM$\uparrow$&LPIPS$\downarrow$&PSNR0$\uparrow$&PSNR$\uparrow$&SSIM$\uparrow$&LPIPS$\downarrow$\\
%    \midrule
%    Nerfacto   &24.99&26.09&0.692&0.336 &22.53&22.94&0.663&0.325\\
%    Mega-NeRF &24.17    &26.02  &0.73 &0.294   &21.70  &22.43 &0.696   &0.282\\
%    InfNeRF-c    &24.62&27.27&0.731&0.274 &22.52&24.02&0.703&0.280\\
%    InfNeRF &\textbf{25.12} &\textbf{27.92} &\textbf{0.781} &\textbf{0.203} &\textbf{23.11}&\textbf{24.70}&\textbf{0.740}&\textbf{0.242}\\
%  \bottomrule
%\end{tabular}
%\end{table*}
\begin{table*}\small
    \setlength{\tabcolsep}{3pt}
  \caption{Quantitative comparison on four public large-scale scene datasets.  We report PSNR0 from full resolution and mean PSNR, SSIM, and LPIPS metrics from 6 resolutions. The \textbf{best results} are highlighted.}
  \label{tab:comparison}
  \begin{tabular}{c|cccc|cccc|cccc|cccc}
    \toprule
    Scene&\multicolumn{4}{c|}{Residential}&\multicolumn{4}{c|}{Sci-Art}&\multicolumn{4}{c|}{Rubble}&\multicolumn{4}{c}{Building}\\
    Metric&PSNR0$\uparrow$&PSNR$\uparrow$&SSIM$\uparrow$&LPIPS$\downarrow$&PSNR0$\uparrow$&PSNR$\uparrow$&SSIM$\uparrow$&LPIPS$\downarrow$&PSNR0$\uparrow$&PSNR$\uparrow$&SSIM$\uparrow$&LPIPS$\downarrow$&PSNR0$\uparrow$&PSNR$\uparrow$&SSIM$\uparrow$&LPIPS$\downarrow$\\
   \midrule
Nerfacto     &24.32  &24.31  &0.772  &0.216  &26.75  &26.48  &0.827  &\textbf{0.182}&24.99&26.09&0.692&0.336 &22.53&22.94&0.663&0.325\\
    PyNeRF &21.30 &24.13 &0.753 &0.299 &23.85 &24.85 &0.815 &0.242 &23.48 &27.08 &0.744 &0.347 &21.70 &\textbf{24.93} &\textbf{0.743} &0.279\\
    Mega-NeRF       &22.52  &22.85  &0.748  &0.265  &24.37  &24.22  &0.808  &0.202 &24.17    &26.02  &0.73 &0.294   &21.70  &22.43 &0.696   &0.282\\
    InfNeRF-c    &23.84  &24.87  &0.785  &0.198  &25.83  &25.81  &0.805  &0.212 &24.62&27.27&0.731&0.274 &22.52&24.02&0.703&0.280\\
    %Ours leaves \\
    InfNeRF   &\textbf{24.66}  &\textbf{25.46}  &\textbf{0.820}  &\textbf{0.158}  &\textbf{27.03}&\textbf{26.96}&\textbf{0.839}&0.192 &\textbf{25.12} &\textbf{27.92} &\textbf{0.781} &\textbf{0.203} &\textbf{23.11}&24.70 &0.740&\textbf{0.242}\\
%    \midrule
%    Scene&\multicolumn{4}{c|}{Rubble}&\multicolumn{4}{c}{Building}\\
     %Metric&PSNR0$\uparrow$&PSNR$\uparrow$&SSIM$\uparrow$&LPIPS$\downarrow$&PSNR0$\uparrow$&PSNR$\uparrow$&SSIM$\uparrow$&LPIPS$\downarrow$\\
%    \midrule
%    Nerfacto   &24.99&26.09&0.692&0.336 &22.53&22.94&0.663&0.325\\
%    Mega-NeRF &24.17    &26.02  &0.73 &0.294   &21.70  &22.43 &0.696   &0.282\\
%    InfNeRF-c    &24.62&27.27&0.731&0.274 &22.52&24.02&0.703&0.280\\
%    InfNeRF &\textbf{25.12} &\textbf{27.92} &\textbf{0.781} &\textbf{0.203} &\textbf{23.11}&\textbf{24.70}&\textbf{0.740}&\textbf{0.242}\\
  \bottomrule
\end{tabular}
\end{table*}

\begin{figure*}
  \centering
  \includegraphics[width=\linewidth]{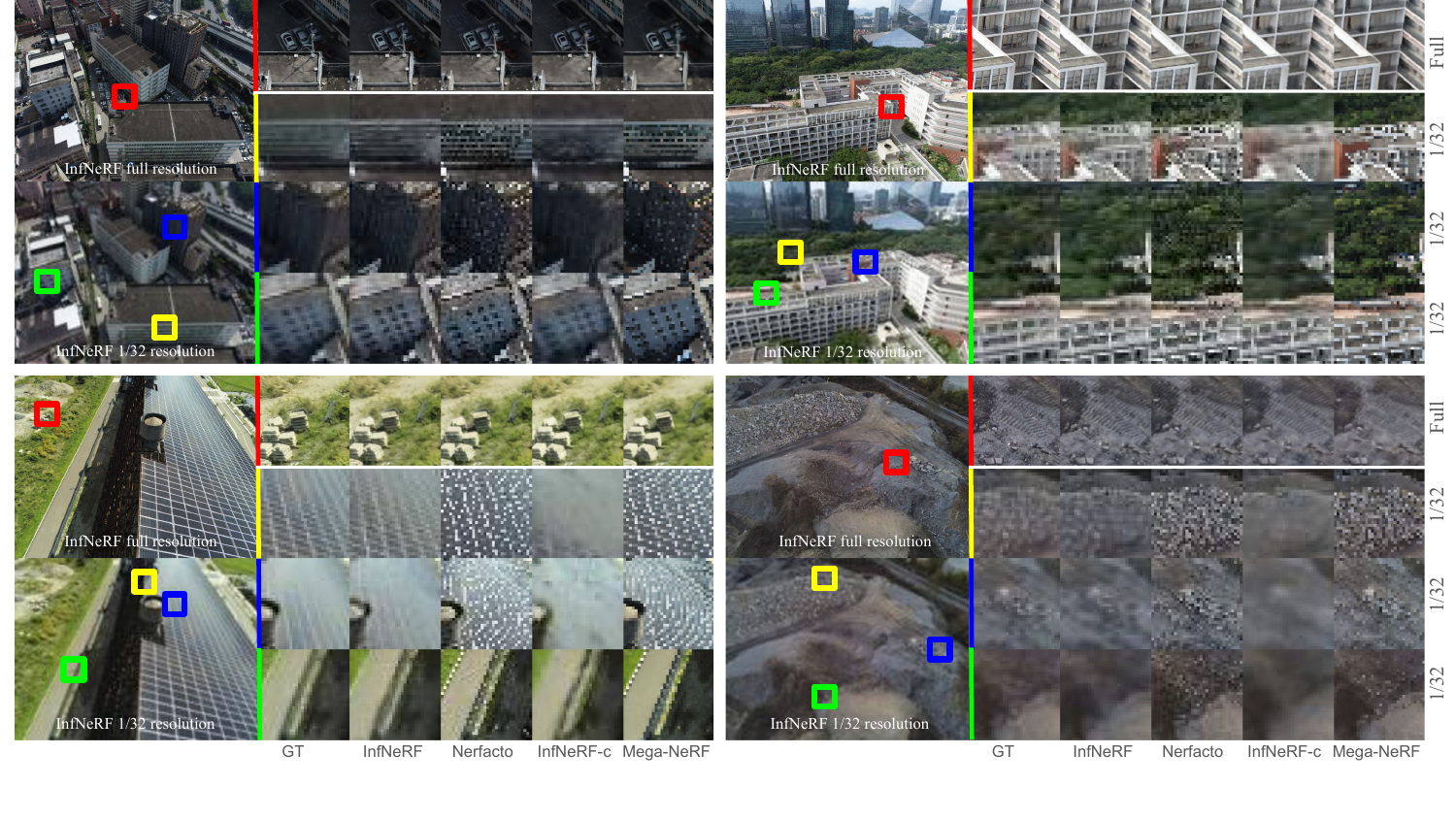}
  \caption{Qualitative comparison between the baseline and InfNeRF on four public large-scale datasets reveals that the reduction of rendering model size do not harm the the rendering quality of InfNeRF at the highest resolution. Meanwhile Mega-NeRF appears blurry and fails to capture details effectively. In low resolution, both Mega-NeRF and Nerfacto suffer severe aliasing, particularly evident in moire patterns on the solar panel. Leveraging the LoD octree, InfNeRF not only captures most details in full resolution but also renders a smooth anti-aliasing image in low resolution. }
  \label{fig:sample}
  \Description{}
\end{figure*}

\begin{figure*}
  \centering
  \includegraphics[width=\linewidth]{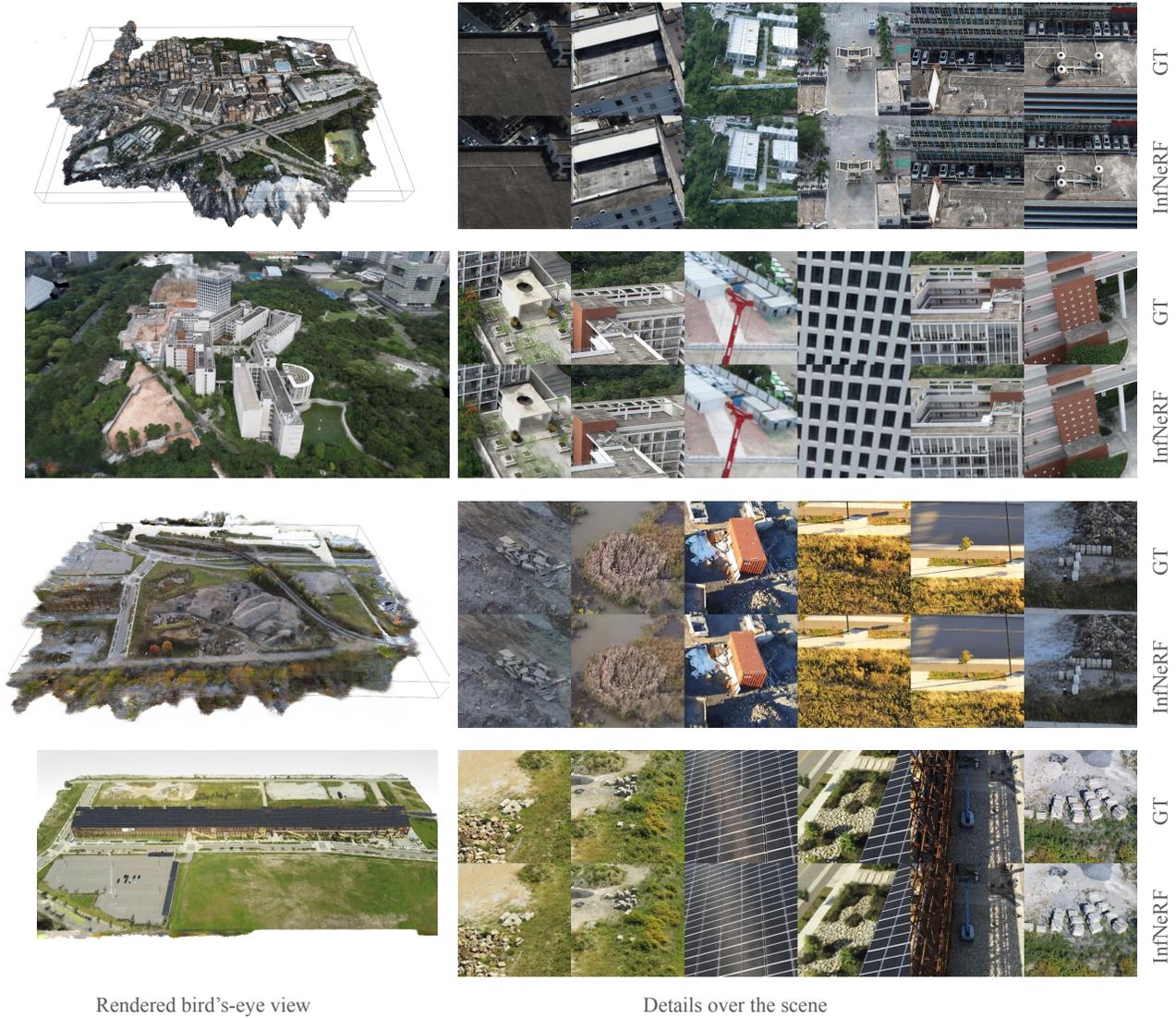}
  \caption{Left: The bird's-eye view showcases four large-scale scenes reconstructed by InfNeRF. Right: The detailed view provides a closer look at certain corner of the scene. InfNeRF demonstrates its ability to reconstruct large-scale scenes while preserving high quality and capturing intricate details. }
  \label{fig:detail}
  \Description{}
\end{figure*}

The PSNR0 (full resolution) and all 3 metrics for multi-resolution are listed in Table \ref{tab:comparison}.
The mean across the four datasets are also plotted in Fig. \ref{fig:2psnr}. 
%Comparing between two methods with similar rendering model sizes, InfNeRF outperforms Mega-NeRF by 2.4dB in multi-resolution PSNR.
%Even the InfNeRF compact with a much smaller rendering model size, still outperforms Mega-NeRF by 1.6 dB.
%This illustrates the overall performance of our proposed method.
When comparing two models with similar model sizes, InfNeRF and Nerfacto, theoretically, they should exhibit similar PSNR0. 
Our results confirm this hypothesis. As depicted along the horizontal axis of Fig. \ref{fig:2psnr}, InfNeRF slightly outperforms Nerfacto by 0.3 dB, demonstrating that InfNeRF's octree partitioning and the limited memory requirement do not compromise the rendering quality.
Furthermore, when considering results of multi-resolution rendering, ploted along the vertical axis, InfNeRF outperforms Nerfacto by 1.31 dB, PyNeRF by 1.01 dB and even outperform Mega-NeRF by 2.4 dB.
This illustrates the  crucial role of the LoD octree in anti-aliasing rendering.
The same conclusion can be drawn from the quality comparison in Fig. \ref{fig:sample}. 
In the last three rows of each dataset, which display the low-resolution images, we can observe that without the anti-aliasing effect, Nerfacto and Mega-NeRF generate noisy results, particularly evident in the heavy moire pattern in the left bottom dataset solar panel region. In contrast, InfNeRF generates satisfactory results at both fine and coarse levels, showcasing its superiority in handling multi-resolution rendering.
More detailed images are provided in Fig.~\ref{fig:detail}.
%We evaluate our method against different methods.
%nfNeRF outperform baseline 
%in full resolution level, Instant-NGP and InNeRF have the same capacity, so has similar performance. about only 0.3db different. For all scale,  Infnerf

%Another interesting observation when compare Our leaf, Single to our Full at level 2 and also mega nerf and our compact at level 3/4. 
%At this two point, the model without anti alias out perform the model with anti alias, because the non anti alias model  render with the whole model trained by all pyramid, while anti alias only render with 1 level. 
%Our compact and Mega NeRF shared same network capacity, in average we still out-performance Mega-NeRF thanks to anti alias result. at level 0 our compact is better then mega-nerf 1.3PSNR because our tree pruning better distribute our network capacity to where it need.
%On average Our full is better then our leaf, and single Instant NGP, except at level 2. because they render with the whole model trained by all level. 
%while we just use one level of the tree.
%The peak shows in level 2 for single INGP and Our leaf, while the peak mega show in level 3, means we even without anti alias effect, our model 
%The high resolution is more difficult to fit, that's why they have a peak at the middle scale. 
%At finest scale 0, even with all our mether outperformance MegaNeRF 

%With limited network capacity Mega-NeRF fail to fit the scale0 well, it only can fix it blurry, that's why when eval on scale 3,4 it have peak and out-performance our model.

\subsection{Experiment On MipNeRF 360 Dataset}
In addition to our previous large-scale experiment, we conduct another experiment with the garden scene of MipNeRF 360 Dataset. This smaller-scale scene allows us to evaluate and compare the performance of various methods. The following methods were included in our comparison:
%We also conduct an experiment on the garden scene of MipNeRF 360 Dataset. The small scale of the scene enale us to compare with more common method, including 
\begin{enumerate*}
\item Gaussian Splatting\cite{kerbl3Dgaussians}.
\item Mip-Splatting\cite{yu2024mip}.
\item Nerfacto~\cite{nerfstudio}.
%\item Mega-NeRF~\cite{turki2022mega} with 25 partitions.
\item PyNeRF~\cite{turki2023pynerf}.
%As this large model takes almost one month to train with 1 GPU, we evaluate the largest and best checkpoint available on their website.
\end{enumerate*} 
All NeRF based methods have similar model size. More details of the experiment setting can be found in the supplementary material.
\begin{table}
\caption{Quantitative comparison on garden scene from MipNeRF 360. We report PSNR at each scale.}
  \label{tab:360}
\begin{tabular}{l|rrrrr}
Resolution                      & $2\times$ & $4\times$ & $8\times$ &$16\times$ & $32\times$ \\ \hline
Gaussian Splatting                         & \cellcolor{Apricot}26.52                 & \cellcolor{Yellow}25.18                 & 21.58                 & 19.01                  & 17.266                 \\
Mip-Splatting &\cellcolor{Salmon}26.87 &\cellcolor{Salmon}27.85 &\cellcolor{Salmon}28.83 &\cellcolor{Apricot}28.69 &\cellcolor{Salmon}27.76\\
Nerfacto              & 23.18                 & 23.5                  & 24.08                 & 23.32                  & 22.09                  \\
PyNeRF                & 23.79                 & 24.76                 & \cellcolor{Yellow} 27.04                 & \cellcolor{Yellow} 27.63                  & \cellcolor{Yellow}24.82                  \\ \hline
InfNeRF               & \cellcolor{Yellow}26.27                 & \cellcolor{Apricot}27.04                 & \cellcolor{Apricot}28.2                  & \cellcolor{Salmon}28.87                  & \cellcolor{Apricot}27.23                  \\ \hline
(1) w/o Trans Loss & 25.63                 & 26.45                 & 27.54                 & 27.92                  & 26.18                  \\
(2) w/o Perturbation      & 25.90                  & 26.87                 & 27.86                 & 28.23                  & 27.20                   \\         
(3) w/o Upper Tree        & 24.71                 & 25.22                 & 24.95                 & 24.02                  & 22.69                  \\
(4) w/o Pyramid           & 26.12                 & 23.04                 & 13.00                 & 8.98                   & 8.52 \\     
(5) w 3 levels &24.78	&25.2	&26.2	&27.26	&25.96
\end{tabular}
%\vspace{-3mm}
\end{table}
As shown in Table. \ref{tab:360}, InfNeRF exhibits competitive outcomes in high-resolution when compared to 3DGS~\cite{kerbl3Dgaussians} and Mip-Splatting~\cite{yu2024mip}. It surpasses PyNeRF and Nerfacto by an margin of 3 dB.
%As the resolution decreases, ZipNeRF, which is specifically tailored for anti-aliasing, demonstrates an improvement. 
As the resolution decreases, 3DGS~\cite{kerbl3Dgaussians} and Nerfacto's performance start to drop. Although anti-aliasing may not be the primary focus of InfNeRF, it consistently delivers commendable results without any additional computational cost.

\subsection{Ablation Study}
\begin{figure}
  \centering
  \includegraphics[width=\linewidth]{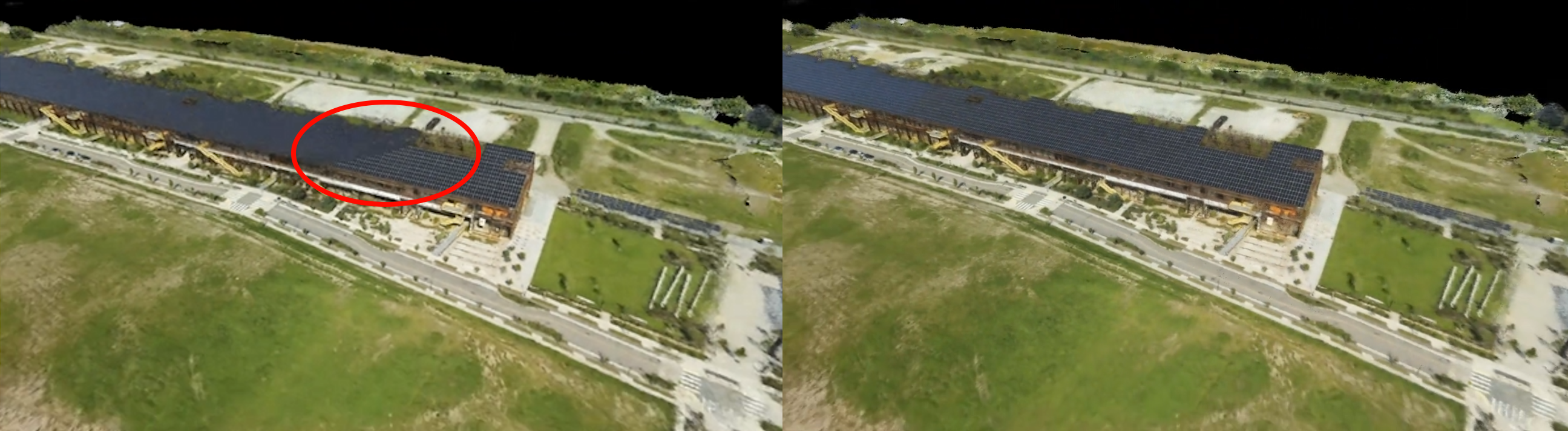}
  \caption{Left: in certain datasets, the absence of sampling sphere perturbation can lead to noticeable discontinuities between different levels of detail. Right: however, by incorporating perturbation in InfNeRF, these transitions are smoothed out, guaranteeing a seamless navigation experience. }
  \label{fig:popup}
  \Description{}
\end{figure}

\begin{figure}
  \centering
  \includegraphics[width=0.8\linewidth]{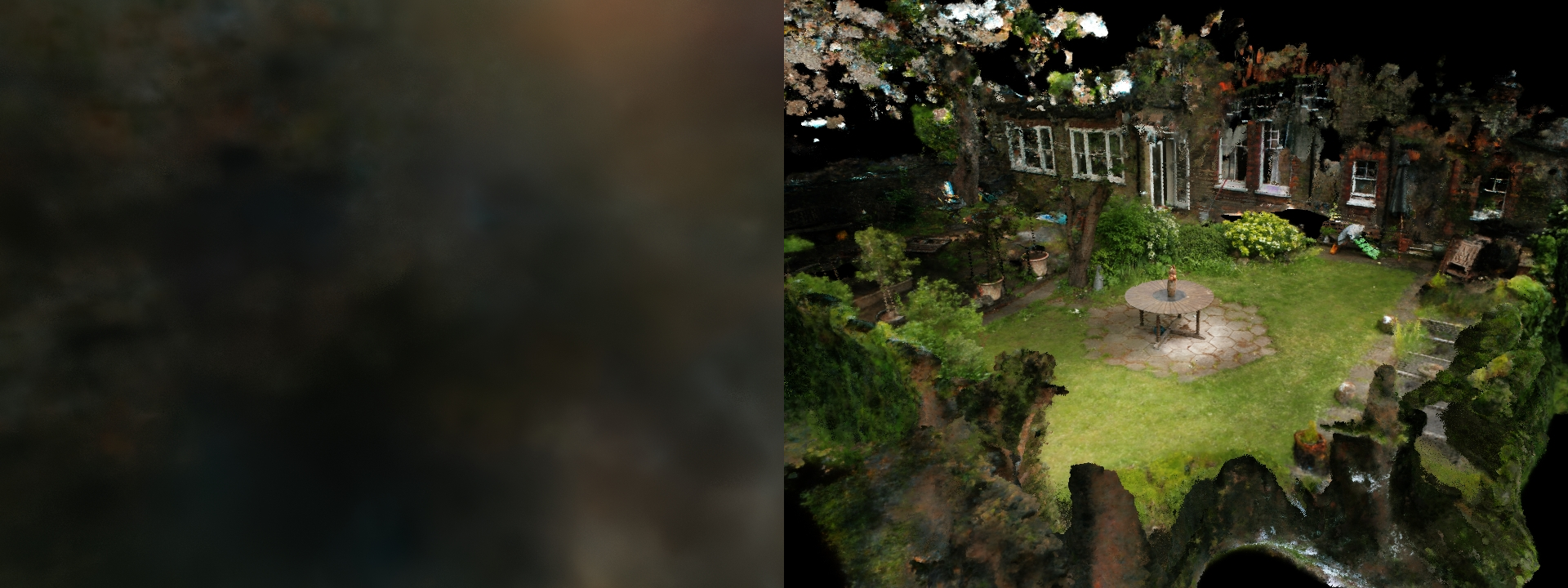}
  \caption{Left: the unobserved sky area is obscured by foggy artifacts, blocking the bird's-eye view. Right: transparency loss can effectively remove this artifact, allowing for a clearer and unobstructed bird's-eye view.}
  \label{fig:fog}
  \Description{}
\end{figure}

Table. \ref{tab:360} also includes an ablation study of our model on garden scene.
\begin{enumerate*}
    \item The transparency loss, improves slightly PSNR, and more importantly, removes the fog blocking the bird eyes view, as shown in Fig. \ref{fig:fog}.
    \item Sampling radius perturbation not only smooth the transition between different LOD levels in some dataset, as depicted in Fig. \ref{fig:popup}, but also improve the overall PSNR.
    \item Training without the upper tree structure, or 
    \item training without image pyramid, results in a significant reduction in PSNR at the low-resolution.
    \item Removing one level of the tree reduces the model size by 4, and decreases the performance by 2 dB. 
    \item Tree pruning reduces the size of the tree to only 19 nodes, which accounted for only 26\% of the full octree.
\end{enumerate*} 

\section{Conclusion and Future work}
In this work, we introduce InfNeRF for large-scale scene reconstruction. 
InfNeRF employs an LoD tree to divide the scene in space and scale into cubes, which are reconstructed using many NeRFs.
During rendering, InfNeRF selectively retrieves a minimal subset of nodes, significantly reducing memory requirements and I/O time for parameter retrieval from disk or cloud storage and also reducing aliasing artifacts. 
In our experiments, InfNeRF achieves superior rendering quality, with an improvement of over 2.4 dB in PSNR while accessing only 17\% of the total parameters.
%Furthermore, InfNeRF imposes no constraints on the underlying NeRF, allowing seamless integration with faster and smaller NeRF models in the future to enhance their scalability and anti-aliasing quality.

InfNeRF demonstrates its potential for large-scale scene reconstruction.
However, there are several aspects for future exploration and improvement. 
Firstly, the reconstruction time and computational burden still exceed traditional, well-optimized photogrammetry methods~\cite{schonberger2016colmap, terra}, necessitating further performance enhancements. 
Secondly, exploring the combination of octree from diverse image sources, such as satellites, planes, and drones, at different times and scales, could enable the creation of larger scenes. 
With these works, digitizing the Earth using NeRF will become a compelling possibility for the future.


\bibliographystyle{ACM-Reference-Format}
\bibliography{sample-base}


\begin{thebibliography}{44}

%%% the Web address.

\ifx \showCODEN    \undefined \def \showCODEN     #1{\unskip}     \fi
\ifx \showDOI      \undefined \def \showDOI       #1{#1}\fi
\ifx \showISBNx    \undefined \def \showISBNx     #1{\unskip}     \fi
\ifx \showISBNxiii \undefined \def \showISBNxiii  #1{\unskip}     \fi
\ifx \showISSN     \undefined \def \showISSN      #1{\unskip}     \fi
\ifx \showLCCN     \undefined \def \showLCCN      #1{\unskip}     \fi
\ifx \shownote     \undefined \def \shownote      #1{#1}          \fi
\ifx \showarticletitle \undefined \def \showarticletitle #1{#1}   \fi
\ifx \showURL      \undefined \def \showURL       {\relax}        \fi
% invisible to TeX
\providecommand\bibfield[2]{#2}
\providecommand\bibinfo[2]{#2}
\providecommand\natexlab[1]{#1}
\providecommand\showeprint[2][]{arXiv:#2}

\bibitem[Barron et~al\mbox{.}(2021)]%
        {barron2021mip}
\bibfield{author}{\bibinfo{person}{Jonathan~T Barron}, \bibinfo{person}{Ben Mildenhall}, \bibinfo{person}{Matthew Tancik}, \bibinfo{person}{Peter Hedman}, \bibinfo{person}{Ricardo Martin-Brualla}, {and} \bibinfo{person}{Pratul~P Srinivasan}.} \bibinfo{year}{2021}\natexlab{}.
\newblock \showarticletitle{Mip-nerf: A multiscale representation for anti-aliasing neural radiance fields}. In \bibinfo{booktitle}{\emph{Proceedings of the IEEE/CVF International Conference on Computer Vision}}. \bibinfo{pages}{5855--5864}.
\newblock


\bibitem[Barron et~al\mbox{.}(2022)]%
        {barron2022mip}
\bibfield{author}{\bibinfo{person}{Jonathan~T Barron}, \bibinfo{person}{Ben Mildenhall}, \bibinfo{person}{Dor Verbin}, \bibinfo{person}{Pratul~P Srinivasan}, {and} \bibinfo{person}{Peter Hedman}.} \bibinfo{year}{2022}\natexlab{}.
\newblock \showarticletitle{Mip-nerf 360: Unbounded anti-aliased neural radiance fields}. In \bibinfo{booktitle}{\emph{Proceedings of the IEEE/CVF Conference on Computer Vision and Pattern Recognition}}. \bibinfo{pages}{5470--5479}.
\newblock


\bibitem[Barron et~al\mbox{.}(2023)]%
        {barron2023zip}
\bibfield{author}{\bibinfo{person}{Jonathan~T Barron}, \bibinfo{person}{Ben Mildenhall}, \bibinfo{person}{Dor Verbin}, \bibinfo{person}{Pratul~P Srinivasan}, {and} \bibinfo{person}{Peter Hedman}.} \bibinfo{year}{2023}\natexlab{}.
\newblock \showarticletitle{Zip-NeRF: Anti-aliased grid-based neural radiance fields}.
\newblock \bibinfo{journal}{\emph{arXiv preprint arXiv:2304.06706}} (\bibinfo{year}{2023}).
\newblock


\bibitem[Burns and Osfield(2004)]%
        {burns2004open}
\bibfield{author}{\bibinfo{person}{Don Burns} {and} \bibinfo{person}{Robert Osfield}.} \bibinfo{year}{2004}\natexlab{}.
\newblock \showarticletitle{Open scene graph a: Introduction, b: Examples and applications}. In \bibinfo{booktitle}{\emph{Virtual Reality Conference, IEEE}}. IEEE Computer Society, \bibinfo{pages}{265--265}.
\newblock


\bibitem[Cernea(2020)]%
        {cernea2020openmvs}
\bibfield{author}{\bibinfo{person}{Dan Cernea}.} \bibinfo{year}{2020}\natexlab{}.
\newblock \showarticletitle{OpenMVS: Multi-view stereo reconstruction library}.
\newblock \bibinfo{journal}{\emph{City}} \bibinfo{volume}{5}, \bibinfo{number}{7} (\bibinfo{year}{2020}).
\newblock


\bibitem[{CesiumJS}(2024)]%
        {web:cesium}
\bibfield{author}{\bibinfo{person}{{CesiumJS}}.} \bibinfo{year}{2024}\natexlab{}.
\newblock \bibinfo{booktitle}{\emph{{CesiumJS Homepage}}}.
\newblock
\urldef\tempurl%
\url{https://cesium.com/}
\showURL{%
Retrieved Jan 20, 2024 from \tempurl}


\bibitem[Chan et~al\mbox{.}(2022)]%
        {Chan2022}
\bibfield{author}{\bibinfo{person}{Eric~R. Chan}, \bibinfo{person}{Connor~Z. Lin}, \bibinfo{person}{Matthew~A. Chan}, \bibinfo{person}{Koki Nagano}, \bibinfo{person}{Boxiao Pan}, \bibinfo{person}{Shalini~De Mello}, \bibinfo{person}{Orazio Gallo}, \bibinfo{person}{Leonidas Guibas}, \bibinfo{person}{Jonathan Tremblay}, \bibinfo{person}{Sameh Khamis}, \bibinfo{person}{Tero Karras}, {and} \bibinfo{person}{Gordon Wetzstein}.} \bibinfo{year}{2022}\natexlab{}.
\newblock \showarticletitle{Efficient Geometry-aware {3D} Generative Adversarial Networks}. In \bibinfo{booktitle}{\emph{CVPR}}.
\newblock


\bibitem[Chen et~al\mbox{.}(2022)]%
        {chen2022tensorf}
\bibfield{author}{\bibinfo{person}{Anpei Chen}, \bibinfo{person}{Zexiang Xu}, \bibinfo{person}{Andreas Geiger}, \bibinfo{person}{Jingyi Yu}, {and} \bibinfo{person}{Hao Su}.} \bibinfo{year}{2022}\natexlab{}.
\newblock \showarticletitle{Tensorf: Tensorial radiance fields}. In \bibinfo{booktitle}{\emph{European Conference on Computer Vision}}. Springer, \bibinfo{pages}{333--350}.
\newblock


\bibitem[Chen et~al\mbox{.}(2023)]%
        {chen2023mobilenerf}
\bibfield{author}{\bibinfo{person}{Zhiqin Chen}, \bibinfo{person}{Thomas Funkhouser}, \bibinfo{person}{Peter Hedman}, {and} \bibinfo{person}{Andrea Tagliasacchi}.} \bibinfo{year}{2023}\natexlab{}.
\newblock \showarticletitle{Mobilenerf: Exploiting the polygon rasterization pipeline for efficient neural field rendering on mobile architectures}. In \bibinfo{booktitle}{\emph{Proceedings of the IEEE/CVF Conference on Computer Vision and Pattern Recognition}}. \bibinfo{pages}{16569--16578}.
\newblock


\bibitem[Clark(1976)]%
        {clark1976hierarchical}
\bibfield{author}{\bibinfo{person}{James~H Clark}.} \bibinfo{year}{1976}\natexlab{}.
\newblock \showarticletitle{Hierarchical geometric models for visible surface algorithms}.
\newblock \bibinfo{journal}{\emph{Commun. ACM}} \bibinfo{volume}{19}, \bibinfo{number}{10} (\bibinfo{year}{1976}), \bibinfo{pages}{547--554}.
\newblock


\bibitem[Daniels et~al\mbox{.}(2008)]%
        {daniels2008quadrilateral}
\bibfield{author}{\bibinfo{person}{Joel Daniels}, \bibinfo{person}{Cl{\'a}udio~T Silva}, \bibinfo{person}{Jason Shepherd}, {and} \bibinfo{person}{Elaine Cohen}.} \bibinfo{year}{2008}\natexlab{}.
\newblock \showarticletitle{Quadrilateral mesh simplification}.
\newblock \bibinfo{journal}{\emph{ACM transactions on graphics (TOG)}} \bibinfo{volume}{27}, \bibinfo{number}{5} (\bibinfo{year}{2008}), \bibinfo{pages}{1--9}.
\newblock


\bibitem[{DJI}(2024)]%
        {terra}
\bibfield{author}{\bibinfo{person}{{DJI}}.} \bibinfo{year}{2024}\natexlab{}.
\newblock \bibinfo{booktitle}{\emph{{Terra}}}.
\newblock
\urldef\tempurl%
\url{https://enterprise.dji.com/dji-terra}
\showURL{%
Retrieved Jan 20, 2024 from \tempurl}


\bibitem[Duckworth et~al\mbox{.}(2023)]%
        {duckworth2023smerf}
\bibfield{author}{\bibinfo{person}{Daniel Duckworth}, \bibinfo{person}{Peter Hedman}, \bibinfo{person}{Christian Reiser}, \bibinfo{person}{Peter Zhizhin}, \bibinfo{person}{Jean-François Thibert}, \bibinfo{person}{Mario Lučić}, \bibinfo{person}{Richard Szeliski}, {and} \bibinfo{person}{Jonathan~T. Barron}.} \bibinfo{year}{2023}\natexlab{}.
\newblock \bibinfo{title}{SMERF: Streamable Memory Efficient Radiance Fields for Real-Time Large-Scene Exploration}.
\newblock
\newblock
\showeprint[arxiv]{2312.07541}~[cs.CV]


\bibitem[Erikson et~al\mbox{.}(2001)]%
        {erikson2001hlods}
\bibfield{author}{\bibinfo{person}{Carl Erikson}, \bibinfo{person}{Dinesh Manocha}, {and} \bibinfo{person}{William~V Baxter~III}.} \bibinfo{year}{2001}\natexlab{}.
\newblock \showarticletitle{HLODs for faster display of large static and dynamic environments}. In \bibinfo{booktitle}{\emph{Proceedings of the 2001 symposium on Interactive 3D graphics}}. \bibinfo{pages}{111--120}.
\newblock


\bibitem[Hoppe(1998)]%
        {hoppe1998smooth}
\bibfield{author}{\bibinfo{person}{Hugues Hoppe}.} \bibinfo{year}{1998}\natexlab{}.
\newblock \showarticletitle{Smooth view-dependent level-of-detail control and its application to terrain rendering}. In \bibinfo{booktitle}{\emph{Proceedings Visualization'98 (Cat. No. 98CB36276)}}. IEEE, \bibinfo{pages}{35--42}.
\newblock


\bibitem[Hoppe(2023)]%
        {hoppe2023progressive}
\bibfield{author}{\bibinfo{person}{Hugues Hoppe}.} \bibinfo{year}{2023}\natexlab{}.
\newblock \showarticletitle{Progressive meshes}.
\newblock In \bibinfo{booktitle}{\emph{Seminal Graphics Papers: Pushing the Boundaries, Volume 2}}. \bibinfo{pages}{111--120}.
\newblock


\bibitem[Hu et~al\mbox{.}(2023b)]%
        {hu2023multiscale}
\bibfield{author}{\bibinfo{person}{Dongting Hu}, \bibinfo{person}{Zhenkai Zhang}, \bibinfo{person}{Tingbo Hou}, \bibinfo{person}{Tongliang Liu}, \bibinfo{person}{Huan Fu}, {and} \bibinfo{person}{Mingming Gong}.} \bibinfo{year}{2023}\natexlab{b}.
\newblock \showarticletitle{Multiscale Representation for Real-Time Anti-Aliasing Neural Rendering}.
\newblock \bibinfo{journal}{\emph{arXiv preprint arXiv:2304.10075}} (\bibinfo{year}{2023}).
\newblock


\bibitem[Hu et~al\mbox{.}(2023a)]%
        {hu2023tri}
\bibfield{author}{\bibinfo{person}{Wenbo Hu}, \bibinfo{person}{Yuling Wang}, \bibinfo{person}{Lin Ma}, \bibinfo{person}{Bangbang Yang}, \bibinfo{person}{Lin Gao}, \bibinfo{person}{Xiao Liu}, {and} \bibinfo{person}{Yuewen Ma}.} \bibinfo{year}{2023}\natexlab{a}.
\newblock \showarticletitle{Tri-miprf: Tri-mip representation for efficient anti-aliasing neural radiance fields}. In \bibinfo{booktitle}{\emph{Proceedings of the IEEE/CVF International Conference on Computer Vision}}. \bibinfo{pages}{19774--19783}.
\newblock


\bibitem[Kerbl et~al\mbox{.}(2023)]%
        {kerbl3Dgaussians}
\bibfield{author}{\bibinfo{person}{Bernhard Kerbl}, \bibinfo{person}{Georgios Kopanas}, \bibinfo{person}{Thomas Leimk{\"u}hler}, {and} \bibinfo{person}{George Drettakis}.} \bibinfo{year}{2023}\natexlab{}.
\newblock \showarticletitle{3D Gaussian Splatting for Real-Time Radiance Field Rendering}.
\newblock \bibinfo{journal}{\emph{ACM Transactions on Graphics}} \bibinfo{volume}{42}, \bibinfo{number}{4} (\bibinfo{date}{July} \bibinfo{year}{2023}).
\newblock
\urldef\tempurl%
\url{https://repo-sam.inria.fr/fungraph/3d-gaussian-splatting/}
\showURL{%
\tempurl}


\bibitem[Kerbl et~al\mbox{.}(2024)]%
        {hierarchicalgaussians24}
\bibfield{author}{\bibinfo{person}{Bernhard Kerbl}, \bibinfo{person}{Andreas Meuleman}, \bibinfo{person}{Georgios Kopanas}, \bibinfo{person}{Michael Wimmer}, \bibinfo{person}{Alexandre Lanvin}, {and} \bibinfo{person}{George Drettakis}.} \bibinfo{year}{2024}\natexlab{}.
\newblock \showarticletitle{A Hierarchical 3D Gaussian Representation for Real-Time Rendering of Very Large Datasets}.
\newblock \bibinfo{journal}{\emph{ACM Transactions on Graphics}} \bibinfo{volume}{43}, \bibinfo{number}{4} (\bibinfo{date}{July} \bibinfo{year}{2024}).
\newblock
\urldef\tempurl%
\url{https://repo-sam.inria.fr/fungraph/hierarchical-3d-gaussians/}
\showURL{%
\tempurl}


\bibitem[Lin et~al\mbox{.}(2024)]%
        {lin2024vastgaussian}
\bibfield{author}{\bibinfo{person}{Jiaqi Lin}, \bibinfo{person}{Zhihao Li}, \bibinfo{person}{Xiao Tang}, \bibinfo{person}{Jianzhuang Liu}, \bibinfo{person}{Shiyong Liu}, \bibinfo{person}{Jiayue Liu}, \bibinfo{person}{Yangdi Lu}, \bibinfo{person}{Xiaofei Wu}, \bibinfo{person}{Songcen Xu}, \bibinfo{person}{Youliang Yan}, {et~al\mbox{.}}} \bibinfo{year}{2024}\natexlab{}.
\newblock \showarticletitle{Vastgaussian: Vast 3d gaussians for large scene reconstruction}. In \bibinfo{booktitle}{\emph{Proceedings of the IEEE/CVF Conference on Computer Vision and Pattern Recognition}}. \bibinfo{pages}{5166--5175}.
\newblock


\bibitem[Lin et~al\mbox{.}(2022)]%
        {UrbanScene3D}
\bibfield{author}{\bibinfo{person}{Liqiang Lin}, \bibinfo{person}{Yilin Liu}, \bibinfo{person}{Yue Hu}, \bibinfo{person}{Xingguang Yan}, \bibinfo{person}{Ke Xie}, {and} \bibinfo{person}{Hui Huang}.} \bibinfo{year}{2022}\natexlab{}.
\newblock \showarticletitle{Capturing, Reconstructing, and Simulating: the UrbanScene3D Dataset}. In \bibinfo{booktitle}{\emph{ECCV}}. \bibinfo{pages}{93--109}.
\newblock


\bibitem[Liu et~al\mbox{.}(2024)]%
        {liu2024citygaussian}
\bibfield{author}{\bibinfo{person}{Yang Liu}, \bibinfo{person}{He Guan}, \bibinfo{person}{Chuanchen Luo}, \bibinfo{person}{Lue Fan}, \bibinfo{person}{Junran Peng}, {and} \bibinfo{person}{Zhaoxiang Zhang}.} \bibinfo{year}{2024}\natexlab{}.
\newblock \showarticletitle{Citygaussian: Real-time high-quality large-scale scene rendering with gaussians}.
\newblock \bibinfo{journal}{\emph{arXiv preprint arXiv:2404.01133}} (\bibinfo{year}{2024}).
\newblock


\bibitem[Luebke(2003)]%
        {luebke2003level}
\bibfield{author}{\bibinfo{person}{David Luebke}.} \bibinfo{year}{2003}\natexlab{}.
\newblock \bibinfo{booktitle}{\emph{Level of detail for 3D graphics}}.
\newblock \bibinfo{publisher}{Morgan Kaufmann}.
\newblock


\bibitem[Martin-Brualla et~al\mbox{.}(2021)]%
        {martinbrualla2020nerfw}
\bibfield{author}{\bibinfo{person}{Ricardo Martin-Brualla}, \bibinfo{person}{Noha Radwan}, \bibinfo{person}{Mehdi S.~M. Sajjadi}, \bibinfo{person}{Jonathan~T. Barron}, \bibinfo{person}{Alexey Dosovitskiy}, {and} \bibinfo{person}{Daniel Duckworth}.} \bibinfo{year}{2021}\natexlab{}.
\newblock \showarticletitle{{NeRF in the Wild: Neural Radiance Fields for Unconstrained Photo Collections}}. In \bibinfo{booktitle}{\emph{CVPR}}.
\newblock


\bibitem[Mildenhall et~al\mbox{.}(2021)]%
        {mildenhall2021nerf}
\bibfield{author}{\bibinfo{person}{Ben Mildenhall}, \bibinfo{person}{Pratul~P Srinivasan}, \bibinfo{person}{Matthew Tancik}, \bibinfo{person}{Jonathan~T Barron}, \bibinfo{person}{Ravi Ramamoorthi}, {and} \bibinfo{person}{Ren Ng}.} \bibinfo{year}{2021}\natexlab{}.
\newblock \showarticletitle{Nerf: Representing scenes as neural radiance fields for view synthesis}.
\newblock \bibinfo{journal}{\emph{Commun. ACM}} \bibinfo{volume}{65}, \bibinfo{number}{1} (\bibinfo{year}{2021}), \bibinfo{pages}{99--106}.
\newblock


\bibitem[M{\"u}ller et~al\mbox{.}(2022)]%
        {muller2022instant}
\bibfield{author}{\bibinfo{person}{Thomas M{\"u}ller}, \bibinfo{person}{Alex Evans}, \bibinfo{person}{Christoph Schied}, {and} \bibinfo{person}{Alexander Keller}.} \bibinfo{year}{2022}\natexlab{}.
\newblock \showarticletitle{Instant neural graphics primitives with a multiresolution hash encoding}.
\newblock \bibinfo{journal}{\emph{ACM Transactions on Graphics (ToG)}} \bibinfo{volume}{41}, \bibinfo{number}{4} (\bibinfo{year}{2022}), \bibinfo{pages}{1--15}.
\newblock


\bibitem[Reiser et~al\mbox{.}(2023)]%
        {reiser2023merf}
\bibfield{author}{\bibinfo{person}{Christian Reiser}, \bibinfo{person}{Rick Szeliski}, \bibinfo{person}{Dor Verbin}, \bibinfo{person}{Pratul Srinivasan}, \bibinfo{person}{Ben Mildenhall}, \bibinfo{person}{Andreas Geiger}, \bibinfo{person}{Jon Barron}, {and} \bibinfo{person}{Peter Hedman}.} \bibinfo{year}{2023}\natexlab{}.
\newblock \showarticletitle{Merf: Memory-efficient radiance fields for real-time view synthesis in unbounded scenes}.
\newblock \bibinfo{journal}{\emph{ACM Transactions on Graphics (TOG)}} \bibinfo{volume}{42}, \bibinfo{number}{4} (\bibinfo{year}{2023}), \bibinfo{pages}{1--12}.
\newblock


\bibitem[Ribelles et~al\mbox{.}(2010)]%
        {ribelles2010improved}
\bibfield{author}{\bibinfo{person}{Jos{\'e} Ribelles}, \bibinfo{person}{Angeles L{\'o}pez}, {and} \bibinfo{person}{Oscar Belmonte}.} \bibinfo{year}{2010}\natexlab{}.
\newblock \showarticletitle{An Improved Discrete Level of Detail Model Through an Incremental Representation.}. In \bibinfo{booktitle}{\emph{TPCG}}. \bibinfo{pages}{59--66}.
\newblock


\bibitem[Schonberger and Frahm(2016)]%
        {schonberger2016colmap}
\bibfield{author}{\bibinfo{person}{Johannes~L Schonberger} {and} \bibinfo{person}{Jan-Michael Frahm}.} \bibinfo{year}{2016}\natexlab{}.
\newblock \showarticletitle{Structure-from-motion revisited}. In \bibinfo{booktitle}{\emph{Proceedings of the IEEE conference on computer vision and pattern recognition}}. \bibinfo{pages}{4104--4113}.
\newblock


\bibitem[Takikawa et~al\mbox{.}(2021)]%
        {takikawa2021nglod}
\bibfield{author}{\bibinfo{person}{Towaki Takikawa}, \bibinfo{person}{Joey Litalien}, \bibinfo{person}{Kangxue Yin}, \bibinfo{person}{Karsten Kreis}, \bibinfo{person}{Charles Loop}, \bibinfo{person}{Derek Nowrouzezahrai}, \bibinfo{person}{Alec Jacobson}, \bibinfo{person}{Morgan McGuire}, {and} \bibinfo{person}{Sanja Fidler}.} \bibinfo{year}{2021}\natexlab{}.
\newblock \showarticletitle{Neural geometric level of detail: Real-time rendering with implicit 3d shapes}. In \bibinfo{booktitle}{\emph{Proceedings of the IEEE/CVF Conference on Computer Vision and Pattern Recognition}}. \bibinfo{pages}{11358--11367}.
\newblock


\bibitem[Tancik et~al\mbox{.}(2022)]%
        {tancik2022block}
\bibfield{author}{\bibinfo{person}{Matthew Tancik}, \bibinfo{person}{Vincent Casser}, \bibinfo{person}{Xinchen Yan}, \bibinfo{person}{Sabeek Pradhan}, \bibinfo{person}{Ben Mildenhall}, \bibinfo{person}{Pratul~P Srinivasan}, \bibinfo{person}{Jonathan~T Barron}, {and} \bibinfo{person}{Henrik Kretzschmar}.} \bibinfo{year}{2022}\natexlab{}.
\newblock \showarticletitle{Block-nerf: Scalable large scene neural view synthesis}. In \bibinfo{booktitle}{\emph{Proceedings of the IEEE/CVF Conference on Computer Vision and Pattern Recognition}}. \bibinfo{pages}{8248--8258}.
\newblock


\bibitem[Tancik et~al\mbox{.}(2023)]%
        {nerfstudio}
\bibfield{author}{\bibinfo{person}{Matthew Tancik}, \bibinfo{person}{Ethan Weber}, \bibinfo{person}{Evonne Ng}, \bibinfo{person}{Ruilong Li}, \bibinfo{person}{Brent Yi}, \bibinfo{person}{Justin Kerr}, \bibinfo{person}{Terrance Wang}, \bibinfo{person}{Alexander Kristoffersen}, \bibinfo{person}{Jake Austin}, \bibinfo{person}{Kamyar Salahi}, \bibinfo{person}{Abhik Ahuja}, \bibinfo{person}{David McAllister}, {and} \bibinfo{person}{Angjoo Kanazawa}.} \bibinfo{year}{2023}\natexlab{}.
\newblock \showarticletitle{Nerfstudio: A Modular Framework for Neural Radiance Field Development}. In \bibinfo{booktitle}{\emph{ACM SIGGRAPH 2023 Conference Proceedings}} \emph{(\bibinfo{series}{SIGGRAPH '23})}.
\newblock


\bibitem[Tanner et~al\mbox{.}(1998)]%
        {tanner1998clipmap}
\bibfield{author}{\bibinfo{person}{Christopher~C Tanner}, \bibinfo{person}{Christopher~J Migdal}, {and} \bibinfo{person}{Michael~T Jones}.} \bibinfo{year}{1998}\natexlab{}.
\newblock \showarticletitle{The clipmap: a virtual mipmap}. In \bibinfo{booktitle}{\emph{Proceedings of the 25th annual conference on Computer graphics and interactive techniques}}. \bibinfo{pages}{151--158}.
\newblock


\bibitem[{Tiled web map}(2024)]%
        {wikitile}
\bibfield{author}{\bibinfo{person}{{Tiled web map}}.} \bibinfo{year}{2024}\natexlab{}.
\newblock \bibinfo{booktitle}{\emph{{Tiled web map--- {W}ikipedia{,} The Free Encyclopedia}}}.
\newblock
\urldef\tempurl%
\url{https://en.wikipedia.org/wiki/Tiled_web_map}
\showURL{%
Retrieved Jan 20, 2024 from \tempurl}


\bibitem[Turki et~al\mbox{.}(2022)]%
        {turki2022mega}
\bibfield{author}{\bibinfo{person}{Haithem Turki}, \bibinfo{person}{Deva Ramanan}, {and} \bibinfo{person}{Mahadev Satyanarayanan}.} \bibinfo{year}{2022}\natexlab{}.
\newblock \showarticletitle{Mega-nerf: Scalable construction of large-scale nerfs for virtual fly-throughs}. In \bibinfo{booktitle}{\emph{Proceedings of the IEEE/CVF Conference on Computer Vision and Pattern Recognition}}. \bibinfo{pages}{12922--12931}.
\newblock


\bibitem[Turki et~al\mbox{.}(2023)]%
        {turki2023pynerf}
\bibfield{author}{\bibinfo{person}{Haithem Turki}, \bibinfo{person}{Michael Zollh{\"o}fer}, \bibinfo{person}{Christian Richardt}, {and} \bibinfo{person}{Deva Ramanan}.} \bibinfo{year}{2023}\natexlab{}.
\newblock \showarticletitle{PyNeRF: Pyramidal Neural Radiance Fields}.
\newblock \bibinfo{journal}{\emph{arXiv preprint arXiv:2312.00252}} (\bibinfo{year}{2023}).
\newblock


\bibitem[Ulrich(2002)]%
        {ulrich2002chunkedLoD}
\bibfield{author}{\bibinfo{person}{Thatcher Ulrich}.} \bibinfo{year}{2002}\natexlab{}.
\newblock \showarticletitle{Rendering massive terrains using chunked level of detail control}. In \bibinfo{booktitle}{\emph{Proc. ACM SIGGRAPH 2002}}.
\newblock


\bibitem[Williams(1983)]%
        {williams1983pyramidal}
\bibfield{author}{\bibinfo{person}{Lance Williams}.} \bibinfo{year}{1983}\natexlab{}.
\newblock \showarticletitle{Pyramidal parametrics}. In \bibinfo{booktitle}{\emph{Proceedings of the 10th annual conference on Computer graphics and interactive techniques}}. \bibinfo{pages}{1--11}.
\newblock


\bibitem[Wu et~al\mbox{.}(2023)]%
        {wu2023scanerf}
\bibfield{author}{\bibinfo{person}{Xiuchao Wu}, \bibinfo{person}{Jiamin Xu}, \bibinfo{person}{Xin Zhang}, \bibinfo{person}{Hujun Bao}, \bibinfo{person}{Qixing Huang}, \bibinfo{person}{Yujun Shen}, \bibinfo{person}{James Tompkin}, {and} \bibinfo{person}{Weiwei Xu}.} \bibinfo{year}{2023}\natexlab{}.
\newblock \showarticletitle{ScaNeRF: Scalable Bundle-Adjusting Neural Radiance Fields for Large-Scale Scene Rendering}.
\newblock \bibinfo{journal}{\emph{ACM Transactions on Graphics (TOG)}} \bibinfo{volume}{42}, \bibinfo{number}{6} (\bibinfo{year}{2023}), \bibinfo{pages}{1--18}.
\newblock


\bibitem[Xiangli et~al\mbox{.}(2022)]%
        {xiangli2022bungeenerf}
\bibfield{author}{\bibinfo{person}{Yuanbo Xiangli}, \bibinfo{person}{Linning Xu}, \bibinfo{person}{Xingang Pan}, \bibinfo{person}{Nanxuan Zhao}, \bibinfo{person}{Anyi Rao}, \bibinfo{person}{Christian Theobalt}, \bibinfo{person}{Bo Dai}, {and} \bibinfo{person}{Dahua Lin}.} \bibinfo{year}{2022}\natexlab{}.
\newblock \showarticletitle{Bungeenerf: Progressive neural radiance field for extreme multi-scale scene rendering}. In \bibinfo{booktitle}{\emph{European conference on computer vision}}. Springer, \bibinfo{pages}{106--122}.
\newblock


\bibitem[Xu et~al\mbox{.}(2023)]%
        {xu2023grid}
\bibfield{author}{\bibinfo{person}{Linning Xu}, \bibinfo{person}{Yuanbo Xiangli}, \bibinfo{person}{Sida Peng}, \bibinfo{person}{Xingang Pan}, \bibinfo{person}{Nanxuan Zhao}, \bibinfo{person}{Christian Theobalt}, \bibinfo{person}{Bo Dai}, {and} \bibinfo{person}{Dahua Lin}.} \bibinfo{year}{2023}\natexlab{}.
\newblock \showarticletitle{Grid-guided Neural Radiance Fields for Large Urban Scenes}. In \bibinfo{booktitle}{\emph{Proceedings of the IEEE/CVF Conference on Computer Vision and Pattern Recognition}}. \bibinfo{pages}{8296--8306}.
\newblock


\bibitem[Yu et~al\mbox{.}(2021)]%
        {yu2021plenoctrees}
\bibfield{author}{\bibinfo{person}{Alex Yu}, \bibinfo{person}{Ruilong Li}, \bibinfo{person}{Matthew Tancik}, \bibinfo{person}{Hao Li}, \bibinfo{person}{Ren Ng}, {and} \bibinfo{person}{Angjoo Kanazawa}.} \bibinfo{year}{2021}\natexlab{}.
\newblock \showarticletitle{Plenoctrees for real-time rendering of neural radiance fields}. In \bibinfo{booktitle}{\emph{Proceedings of the IEEE/CVF International Conference on Computer Vision}}. \bibinfo{pages}{5752--5761}.
\newblock


\bibitem[Yu et~al\mbox{.}(2024)]%
        {yu2024mip}
\bibfield{author}{\bibinfo{person}{Zehao Yu}, \bibinfo{person}{Anpei Chen}, \bibinfo{person}{Binbin Huang}, \bibinfo{person}{Torsten Sattler}, {and} \bibinfo{person}{Andreas Geiger}.} \bibinfo{year}{2024}\natexlab{}.
\newblock \showarticletitle{Mip-splatting: Alias-free 3d gaussian splatting}. In \bibinfo{booktitle}{\emph{Proceedings of the IEEE/CVF Conference on Computer Vision and Pattern Recognition}}. \bibinfo{pages}{19447--19456}.
\newblock


\end{thebibliography}

\end{document}